\documentclass{article}
\usepackage[T1]{fontenc}
\usepackage[utf8]{inputenc}
\pdfoutput=1

\usepackage[utf8]{inputenc}
\usepackage{xcolor}
\usepackage{amssymb}
\usepackage{amsmath}
\usepackage[utf8]{inputenc}
\usepackage{algorithm}
\usepackage[noend]{algpseudocode}
\usepackage{xpatch}
\usepackage[pdftex]{graphicx}
\usepackage{caption}
\usepackage{subcaption}
\usepackage{siunitx}
\usepackage{booktabs}
\usepackage{etoolbox}
\usepackage{hyperref}
\usepackage{authblk}

\makeatletter
\xpatchcmd{\algorithmic}{\itemsep\z@}{\itemsep=1.1ex plus0.1pt}{}{}
\algnewcommand{\LineComment}[1]{\State \(\triangleright\) #1}
\usepackage{geometry}
\date{}
\begin{document}
%
\title{node2coords: Graph Representation Learning with Wasserstein Barycenters}
%
%
%

\author[1]{Effrosyni Simou}
\author[2]{Dorina Thanou}
\author[3]{Pascal Frossard}
\affil[1,3]{Signal Processing Laboratory (LTS4), EPFL}
\affil[2]{Swiss Data Science Center (SDSC), EPFL/ETHZ}

\maketitle

\begin{abstract}
	In order to perform network analysis tasks, representations that capture the most relevant information in the graph structure are needed. However, existing methods do not learn representations that can be interpreted in a straightforward way and that are \color{black}stable \color{black} to perturbations to the graph structure. In this work, we address these two limitations by proposing node2coords, a representation learning algorithm for graphs, which learns simultaneously a low-dimensional space and coordinates for the nodes in that space. The patterns that span the low dimensional space reveal the graph's most important structural information. The coordinates of the nodes reveal the proximity of their local structure to the graph structural patterns. In order to measure this proximity by taking into account the underlying graph, we propose to use Wasserstein distances.  We introduce an autoencoder that employs a linear layer in the encoder and a novel Wasserstein barycentric layer at the decoder. Node connectivity descriptors, that capture the local structure of the nodes, are passed through the encoder to learn the small set of graph structural patterns. In the decoder, the node connectivity descriptors are reconstructed as Wasserstein barycenters of the graph structural patterns. The optimal weights for the barycenter representation of a node's connectivity descriptor correspond to the coordinates of that node in the low-dimensional space. Experimental results demonstrate that the representations learned with node2coords are interpretable, lead to node embeddings that are stable to perturbations of the graph structure and achieve competitive or superior results compared to state-of-the-art methods in node classification.
\end{abstract}

%

\section{Introduction}\label{sec:introduction}
Data is often found to be in the form of a graph structure, like in citation networks or protein-protein interaction networks. Therefore, in order to perform network analysis tasks, for example node classification, it is necessary to design algorithms for the extraction of low-dimensional representations of graphs. This feature extraction process is challenging due to the irregularity of the data structure and has been addressed by employing frameworks from different fields, such as Spectral Graph Theory \cite{chung1997spectral}, Probability Theory \cite{jaynes2003probability} as well as tools from Natural Language Processing \cite{manning1999foundations}, for example. In this work we propose an algorithm for unsupervised learning of graph representations that leverages the mathematical theory of Optimal Transport.

Graph representation learning methods have two important limitations. First, even if their design is justified intuitively, they often do not lead to directly interpretable node embeddings \cite{dalmia2018towards}, \cite{liu2018interpretation}. In particular, there is not a clear interpretation of the values in a node embedding vector. Second, they are highly unstable, as even small perturbations to the graph structure can lead to large changes in the embeddings learned \cite{bojchevski2019certifiable}. 

In this work, we address both limitations by proposing node2coords, an unsupervised learning algorithm that, given a graph structure, learns simultaneously a low-dimensional space as well as node coordinates in that space. Specifically, we propose an autoencoder architecture with a Wasserstein barycentric decoder. The input to the algorithm is given by node connectivity descriptors, which capture the local structure of the nodes. In the encoding step, the node connectivity descriptors pass through a linear layer followed by a softmax activation to obtain a small set of graph patterns. This small set of patterns define the low dimensional space and provide essentially a compressed version of the structural information in the graph. In the decoding step, each node connectivity descriptor is reconstructed as a Wasserstein barycenter of the small set of graph structural patterns that span the low-dimensional space. This step can be thought of as a ``projection" of each node connectivity descriptor in the low dimensional space and it is achieved by learning its optimal barycentric coordinates \cite{bonneel2016wasserstein} in that space. As a result, the value of each feature in the low-dimensional representation of a node can be interpreted as the proximity in terms of Wasserstein distance of the node connectivity descriptor to the corresponding graph structural pattern. 


Experimental results on real and synthetic data demonstrate that the node embeddings obtained with node2coords are stable with respect to small changes of the graph structure. Furthermore, we demonstrate that we achieve competitive or superior results in the tasks of community detection and node classification compared to state-of-the art methods for unsupervised learning of node embeddings that leverage only the graph structure and not node features \cite{chami2020machine}.

The structure of the paper is as follows. In Section \ref{sec:relatedwork} we review the related work in unsupervised learning of graph representations. In Section \ref{sec:preliminaries} we describe the mathematical framework of Optimal Transport that forms the basis of our algorithm. In Section \ref{sec:barycentric_layer} we present in detail our Wasserstein barycenter representation method, which is later proposed as a new layer in the decoder of our graph representation learning algorithm. We outline each block of node2coords in Section \ref{sec:node2barycoords} and explain the interpretations of the graph representations it learns. The experimental evaluation of the performance of node2coords is provided in Section \ref{sec:experiments}. Finally, we conclude and discuss the benefits of node2coords in Section \ref{sec:conclusion}.

\section{Related Work}\label{sec:relatedwork}
The design of algorithms that learn representations for graph structured data has been extensively researched in the last years \cite{cai2018comprehensive}. Such algorithms learn low-dimensional representations for the nodes of the graph that capture the most important information for inference on the graph. If these node embeddings are learned in order to be optimal for a specific downstream prediction task and use labeled data during training, such as the class to which nodes belong for the task of node classification for instance, the corresponding algorithm is a supervised graph representation learning algorithm. Otherwise, if patterns are being learned from the data with no pre-existing labels, it is an unsupervised graph representation learning algorithm. In the case where only the graph structure is being input to the learning algorithm, the node embeddings capture the most relevant structural information in the graph. However, if node features are also available, they can be leveraged by algorithms in order to learn embeddings that capture semantic information as well as the structural one.

We focus here on the methods that learn graph representations in an unsupervised manner and that have been designed to leverage only the graph structure as input, similarly to our algorithm.  Such methods can be grouped into four categories, namely the distance-based methods, matrix factorization methods, skip-gram methods  and autoencoders.

Distance-based embedding methods are algorithms that learn an embedding look-up by forcing nodes that are close in the graph to be mapped as close as possible in the embedding space with respect to a chosen distance metric. Notable methods in this category are Isomap \cite{tenenbaum2000global},  Locally Linear Embedding \cite{roweis2000nonlinear} and Laplacian Eigenmaps \cite{belkin2002laplacian}. These algorithms capture the proximity of nodes on the graph via the geodesic distance, a linear approximation of local neighborhood of nodes and the smoothness of the eigenvectors of the Laplacian matrix of the graph respectively. These three methods measure similarity in the embedding space in terms of Euclidean distance. 

Matrix factorization methods aim to learn low-rank representations of the adjacency matrix of a graph. The first such work is Graph Factorization \cite{ahmed2013distributed} where an efficient algorithm is proposed for the factorization of large graphs. This is followed by HOPE \cite{ou2016asymmetric} which proposes matrix factorization for directed graphs and GraRep \cite{cao2015grarep} which allows for capturing higher order node proximities by leveraging powers of the adjacency matrix.

Skip-gram graph embedding methods are inspired by word embedding methods that use skip-gram \cite{guthrie2006closer} in order to predict context words for a given target word. Thus, by creating sequences of nodes, similarly to sentences of words, it is possible to learn embeddings by maximizing the probability of ``context nodes". Sequences of nodes are generated using random walks and the obtained node sequences are fed to a skip-gram model which maximizes their log-likelihood and provides the node embeddings. In Deepwalk \cite{perozzi2014deepwalk} a neural network is trained by maximizing the probability of predicting ``context nodes" for each  node of the graph. Node2vec \cite{grover2016node2vec} creates  node sequences by generating biased random walks by combining Breadth First Search (BFS) and Depth First Search (DFS) and, as a result, learns node embeddings that capture similarities of nodes in terms of local structure as well as homophily. LINE \cite{tang2015line} learns embeddings that preserve first-order and second-order node proximities. 

Autoencoders for unsupervised learning of graph structures have an encoding and a decoding function that are composed of layers of linear functions and nonlinear activations, thus allowing for the learning of complex graph representations. Generally, the encoder takes as input the adjacency matrix of the graph and maps nodes to their low-dimensional embeddings. The decoder uses the low-dimensional embeddings to reconstruct the input. SDNE \cite{wang2016structural} is an autoencoder, which learns low-dimensional node embeddings that preserve first-order and second-order proximities. In order for this to be achieved, a regularizer that forces nodes that are connected on the graph to be close on the embedding space is added to the graph adjacency matrix reconstruction loss function. DNGR \cite{cao2016deep} creates a similarity matrix from the graph adjacency matrix using random surfing, which is a probabilistic method that employs random walks. The obtained similarity matrix is fed to a denoising autoencoder in order to obtain the node embeddings. 

Our proposed architecture is an autoencoder with a non-linear Wasserstein barycentric layer at the decoder. By introducing this geometry-aware, non-linear operation at the decoder, the graph representations learned have a clear interpretation and lead to stable node embeddings.  \color{black} In \cite{al2020t} Canonical Polyadic decomposition is employed on a multi-view information graph that takes into account the node features in order to learn directly interpretable node embeddings. \color{black} Furthermore, Optimal Transport (OT) was also used for graph representation learning in DVNE \cite{zhu2018deep} where they propose  an autoencoder that maps nodes to Gaussian distributions in order to account for uncertainties in the graph structure and impose neighboring nodes to have a small Wasserstein distance between their corresponding Gaussians. This node embedding method, in terms of architecture, is an autoencoder composed of linear layers at the encoder and the decoder and only employs a Wasserstein-based loss at the objective function. In contrast, in node2coords we propose a new architecture for graph representation learning where the decoder directly employs elements from OT theory.

It is worth noting that OT ideas have been used for representation learning in different contexts. For example, in \cite{schmitz2018wasserstein} the authors propose a dictionary learning algorithm for images where instead of the usual matrix product that determine dictionary and sparse codes, they propose to use Wasserstein barycenters for the image reconstruction. In \cite{yurochkin2019hierarchical} the authors propose a hierarchical optimal transport algorithm where they model documents as distributions over topics and use optimal transport distances to compare documents on the smaller topic space. In our work, we propose an autoencoder for graph representation learning that incorporates a novel Wasserstein barycentric layer at the decoder, which allows for geometry aware, non-linear combinations of graph patterns. \color{black} Finally, we would like to mention that OT has been used recently to solve other problems related to graph structured data. For instance, in \cite{maretic2019got} OT was leveraged for graph comparison and in \cite{chen2020graph} in order to solve the problem of graph alignment.\color{black}

\section{Preliminaries}\label{sec:preliminaries}
\subsection{Optimal Transport}
Optimal Transport \cite{villani2008optimal} is a mathematical theory that allows for the definition of geometry-aware distances between probability distributions. These distances take into account the geometry of the space on which the distributions are defined by leveraging a cost that captures the distances on that space. 

The optimal transportation problem was first formulated by Monge \cite{monge1781memoire} in order to find the optimal way to transport the entirety of the mass of a pile of sand to a different position. This problem was later relaxed by Kantorovich by allowing for split of mass during the transportation. The Kantorovich OT problem \cite{kantorovich1942translation} aims to find a coupling $\Gamma$, where $\Gamma(x,y)$ describes the probability of moving mass from $x$ to $y$. In the discrete case where mass can only be found at specific positions, histograms can be considered and the geometry-aware cost is an $n \times m$ matrix $C$, where $n$ is the dimensionality of the source histogram $I_0$ and $m$ is the dimensionality of the target histogram $I_1$. The Kantorovich formulation of the OT problem is as follows:
\begin{equation}\label{eq:KantorovichHistograms}
K(I_0,I_1) = \min_{\Gamma \in \mathcal{U}(I_0,I_1)} \langle C, \Gamma \rangle
\end{equation}
where:
\begin{equation}\label{eq:KantorovichHistogramsMassPreservation}
\mathcal{U}(I_0,I_1) = \{ \Gamma \in \mathbf{R}_+^{n \times m}: \Gamma 1_m=I_0  \text{ and } \Gamma^{\top}1_n= I_1 \}
\end{equation}
is the polytope of transportation couplings that satisfy the mass preservation constraints and $1_n$, $1_m$ are vectors of ones with dimension $n$ and $m$ respectively. The mass preservation constraints guarantee that the entirety of mass of the source $I_0$ is transported to the target $I_1$.

The $p$-Wasserstein distance is defined as a specific case of the OT problem when $n=m$, and the cost matrix $C$ in Eq. (\ref{eq:KantorovichHistograms}) can be expressed as the $p$-th elementwise power of the distance matrix $D$ of the $n$-dimensional space. The $p$-Wasserstein distance is therefore obtained by:
\begin{equation}\label{eq:WassersteinDistanceDefinition}
W_p(I_0, I_1)=\left (\min_{\Gamma \in \mathcal{U}(I_0,I_1)} \langle D^p, \Gamma \rangle \right )^{1/p}.
\end{equation} 

\subsection{Entropy Regularization}\label{ssec:sinkh}
The problem in Eq. (\ref{eq:WassersteinDistanceDefinition}), (\ref{eq:KantorovichHistogramsMassPreservation}) is a linear program and therefore can be solved with LP solvers such as the simplex algorithm \cite{goldfarb1977practicable}. In \cite{cuturi2013sinkhorn} the problem is regularized with the negative entropy of the coupling $\Gamma$ and an efficient algorithm is proposed for its solution. The entropy-regularized Wasserstein distance is defined as:
\begin{equation}\label{eq:EntropyRegularizedWassersteinDistanceDefinition}
\mathcal{W}_p^{\epsilon}(I_0, I_1) = \min_{\Gamma \in \mathcal{U}(I_0,I_1)} \langle C, \Gamma \rangle - \epsilon H(\Gamma) 
\end{equation} 
where $H(\Gamma)=-\mathop{\sum_{i=1}^{n}\sum_{j=1}^{n}}\Gamma(i,j)\log(\Gamma(i,j)-1)$ is the entropy of the transportation coupling and $\epsilon$ is the regularization parameter. This regularized problem can be solved efficiently with matrix scaling of the so-called Gibbs kernel $K=e^{-\frac{C}{\epsilon}}$. The Gibbs kernel is geometry aware, since it is a function of the cost $C$, but it is also a non-negative matrix.  The non-negativity of $K$ leads to the efficient solution of the problem in Eq. (\ref{eq:EntropyRegularizedWassersteinDistanceDefinition}), (\ref{eq:KantorovichHistogramsMassPreservation}) using Sinkhorn's matrix scaling algorithm \cite{knight2008sinkhorn}. Instead of the coupling $\Gamma$, the parameters that are being optimized in that case are the scaling vectors $u$ and $v$, from which the coupling $\Gamma$ can be eventually obtained as $\Gamma= \operatorname{diag}(u)K \operatorname{diag}(v)$. The solution of the problem in Eq. (\ref{eq:EntropyRegularizedWassersteinDistanceDefinition}), (\ref{eq:KantorovichHistogramsMassPreservation}) with $L$ Sinkhorn iterations is shown in Algorithm \ref{algo:SinkhornIterations}.

\begin{algorithm}
	\caption{Sinkhorn Iterations for Wasserstein Distance}
	\begin{algorithmic}
		\State{\bf{Input: }}$ I_0, I_1$
		\LineComment{Initialization}
		\State $u^{(0)}=1_N $ 
		\For{$l \gets 0$ to $L-1$}
		\LineComment{Update first scaling vector}
		\State $v^{(l)}=\displaystyle\frac{I_0}{K^{\top}u^{(l)}}$
		\LineComment{Update second scaling vector}
		\State $u^{(l+1)}=\displaystyle\frac{I_1}{Kv^{(l)}}$
		\EndFor
		\State $\Gamma^{(L)}=\operatorname{diag}(u^{(L)})K \operatorname{diag}(v^{(L-1)})$
		\State \textbf{return} $W_p^{\epsilon}(I_0, I_1)=\langle C, \Gamma^{(L)} \rangle $   
	\end{algorithmic} \label{algo:SinkhornIterations}
\end{algorithm}


In some cases it may be desired to relax the mass preservation constraints in order to compare a source $I_0$ and a target $I_1$ that do not have the same mass. In \cite{chizat2018scaling}, it is proposed to control the mass variation through the parameter $\rho$, leading to the definition of the unbalanced Wasserstein distance with entropy regularization:

\begin{flalign}\label{eq:EntropyRegularizedUnbalancedWassersteinDistanceDefinition}
\begin{split}
\mathcal{W}_{p}^{\epsilon,\rho}(I_0,I_1)= & \operatorname*{min}_{\Gamma \in \mathbf{R}^{N\times N}_{+}}\langle C, \Gamma \rangle + \epsilon H(\Gamma) \\
& + \rho (\operatorname{KL}(\Gamma 1_N|I_0)+ \operatorname{KL}(\Gamma^{\top}1_N|I_1)),
\end{split}
\end{flalign}
where $\operatorname{KL}(\cdot | \cdot)$ is the Kullback-Leibler divergence \cite{kullback1951information}. This problem 
can also be solved with Sinkhorn iterations \cite{chizat2018scaling}, equivalent to those of Algorithm \ref{algo:SinkhornIterations}, but the scaling vectors are raised to the power $\frac{\rho}{\rho + \epsilon}$.

\subsection{Wasserstein Barycenters}
Given the Wasserstein distance, in \cite{agueh2011barycenters} the notion of a Wasserstein barycenter of a set of histograms is introduced. The Wasserstein barycenter is an interpolation of $S$ histograms $\{I_k\}_{k=1}^{S}$ with weights $\{\lambda_k\}_{k=1}^S$ and it is defined as:

\begin{flalign}\label{eq:WassersteinBarycenter}
\begin{split}
&\hat{b} = \operatorname*{argmin}_{b}\sum_{k=1}^{S} \lambda_k W_{p}(I_k,b) \\
&\text{subject to }\sum_{k=1}^{S}\lambda_k = 1.
\end{split}
\end{flalign}

The histogram $\hat{b}$ is called the Wasserstein barycenter and the weights $\{\lambda_k\}_{k=1}^S$ are referred to as barycentric coordinates. In the specific case where the Wasserstein distance employed is the unbalanced Wasserstein distance of Eq. (\ref{eq:EntropyRegularizedUnbalancedWassersteinDistanceDefinition})  the obtained barycenter is the unbalanced Wasserstein barycenter. 

When computing the barycenter of $S$ histograms we are solving simultaneously $S$ optimal transport problems between each of the $S$ known targets, which are the $S$ histograms $\{I_k\}_{k=1}^{S}$, and the unknown source, which is the barycenter $b$. Therefore, for the entropy-regularized case described above, $S$ sets of scaling vectors $u$ and $v$ have to be computed. The computation of the unbalanced Wasserstein barycenter can be performed through Sinkhorn iterations \cite{janati2018wasserstein} as discussed in Section \ref{ssec:sinkh}. An extra step is added for the estimation of the unknown barycenter $b$, which is needed for the update of the second scaling vectors \cite{benamou2015iterative}. The Sinkhorn iterations for the computation of the unbalanced Wasserstein barycenter are shown in Algorithm \ref{algo:unbalanced}.

\begin{algorithm}[!ht]
	\caption{Sinkhorn Iterations for Unbalanced Wasserstein Barycenter}
	\begin{algorithmic}
		\State{\bf{Input: }}$ \{I_k\}_{k=1}^S, \{\lambda_k\}_{k=1}^S$
		\LineComment{Initialization}
		\For{$k \gets 1$ to $S$}
		\State $u_k^{(0)}=1_N $ 
		\EndFor
		\For{$l \gets 0$ to $L-1$}
		\LineComment{Update first scaling vectors}
		\For{$k \gets 1$ to $S$}
		\State $v_k^{(l)}=\left(\displaystyle\frac{I_k}{K^{\top}u_k^{(l)}} \right)^{\frac{\rho}{\rho + \epsilon}}$
		\EndFor  
		\LineComment{Estimate Barycenter}
		\State $b^{(l)}=\left(\sum\limits_{k=1}^S \lambda_k (u_k^{(l)}\odot Kv_k^{(l)})^{\frac{\epsilon}{\epsilon + \rho}}\right)^{\frac{\epsilon+\rho}{\epsilon}}$
		\LineComment{Update second scaling vectors}
		\For{$k \gets 1$ to $S$}
		\State $u_k^{(l+1)}=\left(\displaystyle\frac{b^{(l)}}{Kv_k^{(l)}}\right)^{\frac{\rho}{\rho + \epsilon}}$
		\EndFor  
		\EndFor
		\State \textbf{return} $\hat{b}=b^{(L-1)}$   
	\end{algorithmic} \label{algo:unbalanced}
\end{algorithm}


\section{Wasserstein Barycenters for Graph Representation Learning}\label{sec:barycentric_layer}
\color{black}In this section we propose a Wasserstein barycenter representation method for graphs, which provides differentiable, geometry-aware non-linearities and can be incorporated in different deep network architectures for graph representation learning. For instance, it can be used to learn the  optimal Wasserstein barycenter representations of its inputs by learning the barycentric weights. Alternatively, it can be used to aggregate multiple feature maps into one, using fixed barycentric weights, in a Graph Convolutional Network \cite{NIPS2016_6081}. \color{black}  This new method is also employed in the barycentric layer of node2coords.

\subsection{Wasserstein Barycenter of Graph Patterns}\label{ssec:bary_intui}
Given a graph $G$ of $N$ nodes with adjacency matrix $\mathcal{A}$, the input to our method can be any set of non-negative, $N$-dimensional vectors with unit $l_1$ norm. We refer to such vectors, input to our proposed graph Wasserstein barycenter computation method, as \emph{graph patterns}. \color{black}
We propose to compute Wasserstein barycenters for graph patterns by taking into account the underlying graph geometry through the diffusion distance $D_{\tau}$ \cite{coifman2006diffusion}. Hence, the geometry aware cost $C$ is chosen to be $C=D_{\tau}^{p=1}$ and the Gibbs kernel becomes $K=e^{-\frac{D_{\tau}}{\epsilon}}$. The diffusion distance $D_{\tau}$ captures the similarity of node connections in $\tau$ hops and is computed using the $\tau$-th power of a Markov matrix $P$ defining a random walk on the graph. The degree of node $i$ is defined as:
\begin{equation}
d(i)= \sum_{j=1}^{N}\mathcal{A}(i, j)
\end{equation}
and the Markov matrix $P$ as:
\begin{equation}
P(i, j) = \frac{\mathcal{A}(i, j)}{d(i)}\cdot
\end{equation}

The diffusion distance $D_{\tau}$ between a pair of nodes $i, j$ is computed as:
\begin{align*}
D^2_{\tau}(i,j)=\|P^{\tau}(i, \cdot) - P^{\tau}(j, \cdot)\|_{L^2}^2\\
=\sum_{u=1}^N\frac{(P^{\tau}(i, u)-P^{\tau}(j, u))^2}{\pi(u)}
\end{align*}
where:
\begin{equation}
\pi(x)=\frac{d(x)}{\sum_{y=1}^N d(y)}\cdot
\end{equation}

Given the Gibbs kernel $K=e^{-\frac{D_{\tau}}{\epsilon}}$, we can compute unbalanced Wasserstein barycenters with Sinkhorn iterations as explained in Section III.

\begin{figure}[!h]
	\centering
	\hspace{-2mm}
	\includegraphics[width=0.7\textwidth]{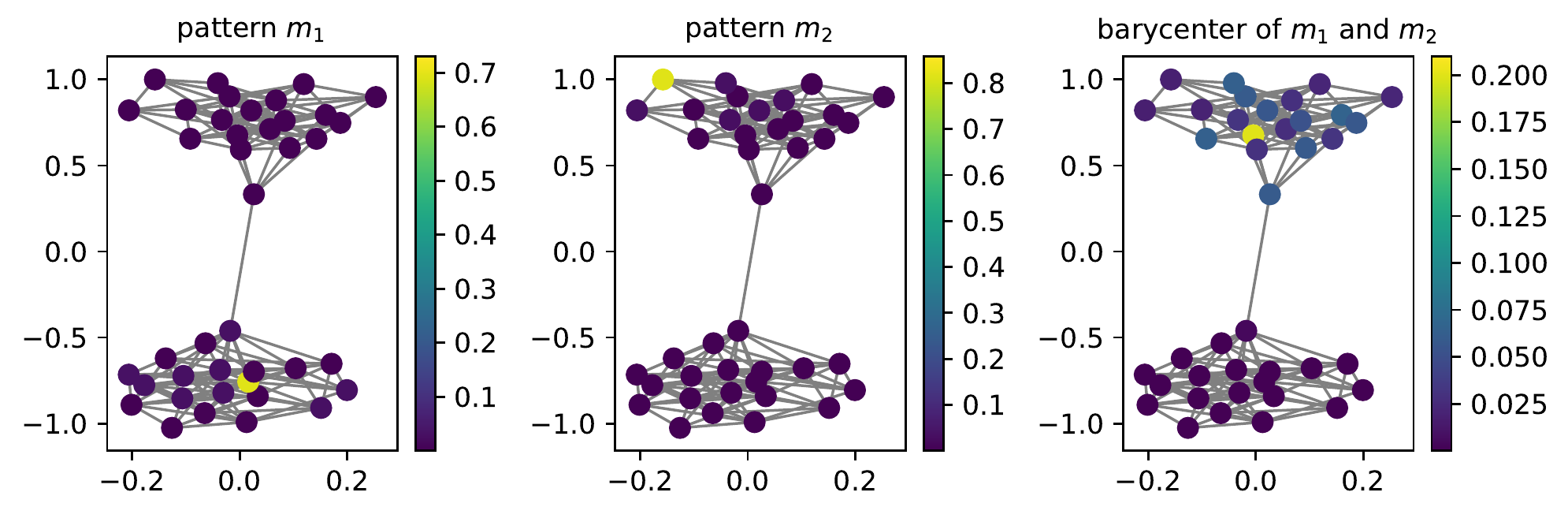}
	\caption{Localized graph patterns $m_1$, $m_2$ plotted on the graph and their Wasserstein barycenter $b$ for $\lambda_1=0.2$ and $\lambda_2=0.8$. $m_1$, $m_2$ and $b$ are plotted on the graph in order to highlight that the barycentric interpolation takes into account the underlying graph.}\label{fig:wassbary}
\end{figure}

We now provide an example that illustrates the representations obtained with Wasserstein barycenters of graph patterns. 
Consider a graph composed of two clusters. Let $m_1$ and $m_2$ be two graph patterns localized at each cluster of the graph. The graph patterns $m_1$, $m_2$ as well as their unbalanced Wasserstein barycenter $b$ for $\lambda_1=0.2$ and $\lambda_2=0.8$ are shown in Fig. (\ref{fig:wassbary}).  \color{black} We note that $m_1$ and $m_2$ are $N$-dimensional graph patterns. Their barycenter $b$ is an interpolation that takes into account the graph through the diffusion distance cost $C=D_{\tau}$. In Fig. (\ref{fig:wassbary}) we plot $m_1$, $m_2$ and $b$ on the graph in order to highlight that the Wasserstein barycenter $b$ is a geometry-aware, non-linear interpolation of $m_1$ and $m_2$. \color{black} It can be seen that the values of the barycentric coordinates $\lambda_1$, $\lambda_2$ quantify the proximity of the barycenter $b$ with respect to the patterns $m_1$, $m_2$. 

Also, it can be seen that the barycenter $b$ has a larger support than the graph patterns $m_1$ and $m_2$ that are being interpolated because of the entropy regularization of Eq. (\ref{eq:EntropyRegularizedUnbalancedWassersteinDistanceDefinition}). Specifically, as the value of the entropy regularization parameter $\epsilon$ becomes larger, the barycenter tends to be uniform over the graph.  

The graph patterns that can be interpolated with Wasserstein barycenters are not restricted to be localized, as those shown in Fig. (\ref{fig:wassbary}). Our proposed method can be integrated in different algorithms for learning representations of graph structured data by providing geometry aware, non-linear interpolations. In the next Section, we integrate it in an autoencoder architecture for unsupervised graph representation learning.\color{black} 


\subsection{Efficient Wasserstein Barycenter Computation Method}\label{ssec:efficient}
\color{black}
We now propose our efficient implementation for computing Wasserstein barycenters of graph patterns. Our method takes as input a matrix of $S$ graph patterns $M_S=[m_1, \ldots, m_S]$, with $m_i$ an $N \times 1$ graph pattern, and outputs their $J$ Wasserstein barycenters $B_J=[b_1, \ldots, b_J]$, as computed for $J$ sets of barycentric coordinates $\Lambda$ given by:
\begin{equation}
\Lambda= 
\left [
\begin{array}[tb]{*{3}{c}}
\lambda_{1,1} & \ldots &  \lambda_{1, S} \\
\multicolumn{3}{c}{\dotfill}  \\
\lambda_{J,1} & \ldots & \lambda_{J, S}  \\
\end{array}  
\right ]\cdot
\end{equation}

The barycenter $b_i$ is an $N$-dimensional vector obtained as the barycenter of the graph patterns in $M_S$ with weights $\Lambda(i, \cdot)=[\lambda_{i,1}, \ldots, \lambda_{i,S}]$.

As explained in Section \ref{sec:preliminaries}, at each Sinkhorn iteration it is needed to update each of the $S$ scaling vectors $v$ and then, once the barycenter has been estimated, update each of the $S$ scaling vectors $u$. In the case where the updates of the $S$ sets of scaling vectors are performed serially, the computation of the barycenter is inefficient in terms of time complexity. 

We avoid this increase in the time complexity by proposing a parallelized with respect to $S$, and computationally efficient method for the barycenter computation. Specifically, we update in parallel the $S$ scaling vectors $v$. Then, after the barycenter estimation, we also update in parallel the $S$ scaling vectors $u$. Our proposal for the efficient Wasserstein barycenter computation is demonstrated in Algorithm \ref{algo:BarycenterLayer}. An important element of the parallelization is due to the fact that the matrix-vector multiplications of the Gibbs kernel $K$ with the scaling vectors $u_k$ can be implemented in parallel as a matrix-matrix multiplication of the $N \times N$ matrix $K$ with the $N \times S$ matrix $U=[u_1, \ldots, u_S]$, whose columns are the $S$ scaling vectors $\{u_k\}_{k=1}^S$. Similarly, the matrix-vector multiplications of the matrix $K$ with the $S$ scaling vectors $\{v_k\}_{k=1}^S$  can be implemented in parallel as a matrix-matrix multiplication of $K$ with $V=[v_1, \ldots, v_S]$. We demonstrate step by step the computations of our proposed method and analyse their time complexity. 

The first step is the update of the scaling vectors in $V$. In our implementation, the update of the $S$ scaling vectors $V$ is equivalent to the matrix multiplication $K^{\top}U$, the element-wise division of the $N \times S$ matrices $M_S$ and $K^{\top}U$ and the elementwise exponentiation of the resulting $N \times S$ matrix to $\frac{\rho}{\rho + \epsilon}$. The time complexity for the update of the scaling vectors $V$ is therefore $\mathcal{O}(N^2S+2NS)$.

The second step is the estimation of the barycenter. We perform this update efficiently using matrix operations as:
\begin{equation}
B_J^{(l)}(i, \cdot)=(((1_N \otimes \Lambda(i, \cdot)) \odot (U \odot KV)^{\frac{\epsilon}{\epsilon + \rho}})1_S)^{\frac{\epsilon+\rho}{\epsilon}}.
\end{equation}
The time complexity of this operation is $\mathcal{O}(N^2S+5NS+N)$.

The third and final step is the update of the $S$ scaling vectors $U$. This step is equivalent to the matrix multiplication $KV$, the element-wise division of the $N \times S$ matrices $B_J(i, \cdot) \otimes 1_S$ and $KV$ and the elementwise exponentiation of the resulting $N \times S$ matrix to $\frac{\rho}{\rho + \epsilon}$. Thus, the time complexity for the update of the scaling vectors $U$ is $\mathcal{O}(N^2S+3NS)$.

\begin{algorithm}
	\caption{Barycenter Computation in node2coords}
	\begin{algorithmic}
		\State{\bf{Input: }}$ M_S, \Lambda(i, \cdot)$
		\LineComment{Initialization}
		\State $U^{(0)}=1_{N \times S}$ 
		\For{$l \gets 0$ to $L-1$}
		\LineComment{Update first scaling vectors}
		\State $V^{(l)}=\left(\displaystyle\frac{M_S}{K^{\top}U^{(l)}} \right)^{\frac{\rho}{\rho + \epsilon}}$  
		\LineComment{Estimate Barycenter}
		\State $B_J^{(l)}(i, \cdot)=\left(\left((1_N \otimes \Lambda(i, \cdot)) \odot (U \odot KV)^{\frac{\epsilon}{\epsilon + \rho}}\right)1_S\right)^{\frac{\epsilon+\rho}{\epsilon}}$
		\LineComment{Update second scaling vectors}
		\State $U^{(l+1)}=\left(\displaystyle\frac{B_J^{(l)}(i, \cdot) \otimes 1_S}{KV^{(l)}}\right)^{\frac{\rho}{\rho + \epsilon}}$
		\EndFor
		\State \textbf{return} $B_J(i, \cdot)^{(L-1)}$
		
	\end{algorithmic} \label{algo:BarycenterLayer}
\end{algorithm}

As a result, the time complexity for our implementation of a Sinkhorn iteration of the barycenter $B_J(i, \cdot)$ is $\mathcal{O}(3N^2S+10NS+N)$, which means that it scales quadratically with respect to the number of nodes $\mathcal{O}(N^2)$. Finally, we note that our implementation of the Wasserstein barycenter computation allows for the $J$ barycenters in $B_J$ to be computed simultaneously using broadcasting operations, which are common in libraries such as PyTorch  \cite{paszke2017automatic}. As a result, the complexity of each Sinkhorn iteration for the computation in parallel of $J$ barycenters is $\mathcal{O}(JN^2S)$. Therefore, the overall complexity of the barycenter computation, which is composed of $L$ Sinkhorn iterations, is $\mathcal{O}(LJN^2S)$. The number of Sinkhorn iterations $L$ needed in order for the barycenter computation to converge increases as the entropy regularization parameter $\epsilon$ decreases \cite{kroshnin2019complexity}.

\begin{figure*}[t!]
	\centering
	\includegraphics[width=0.9\textwidth]{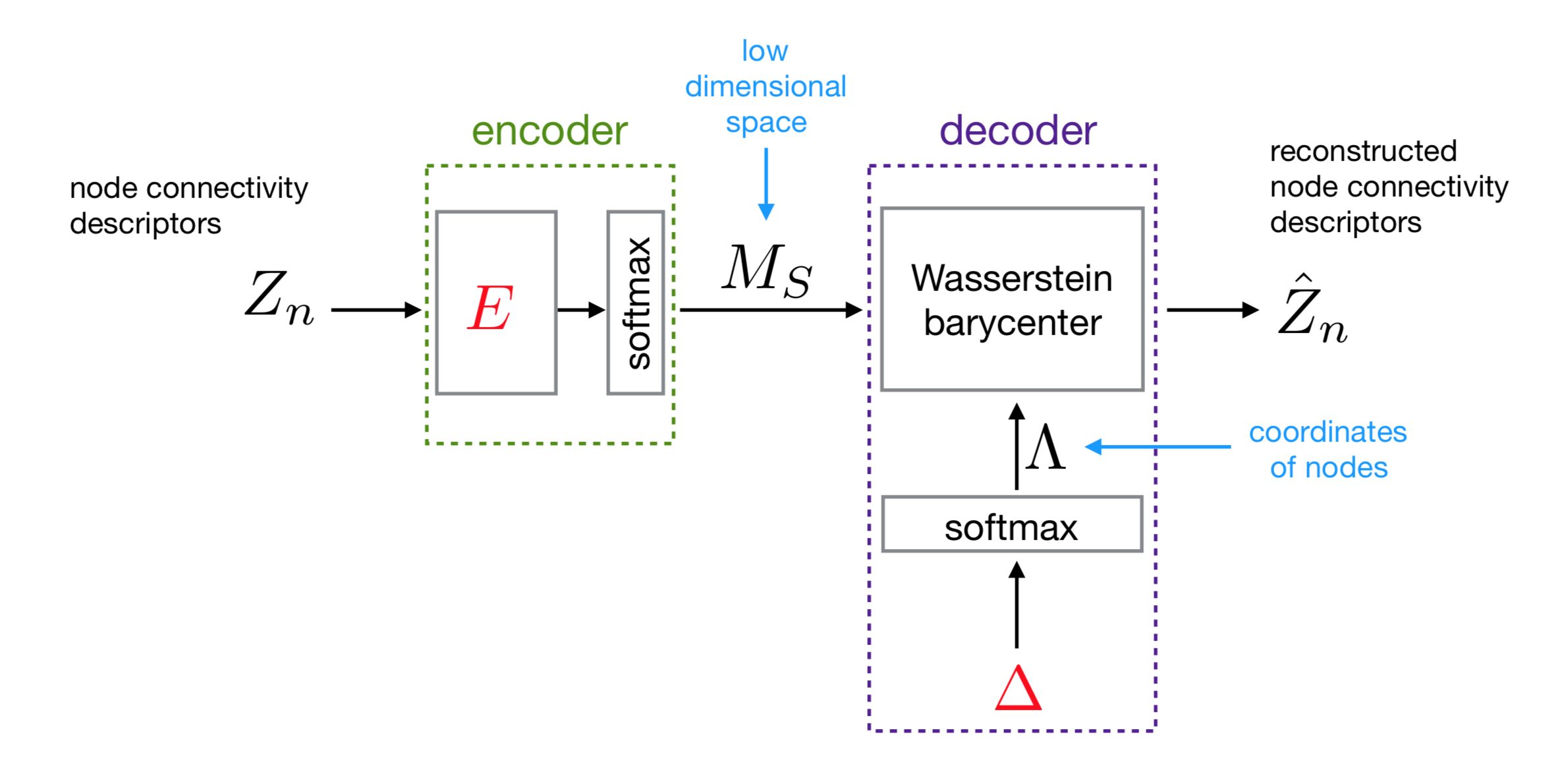}
	\caption{Node2coords block scheme. In the encoder, the node connectivity descriptors are passed through a linear layer followed by a softmax activation to obtain the small set of graph structural patterns that define the low dimensional space $M_S$. In the decoder, the node connectivity descriptors are reconstructed as Wasserstein barycenters of the patterns in $M_S$ by optimizing for their barycentric coordinates $\Lambda$. The barycentric coordinates are re-parametrized through a softmax layer in order to guarantee that they sum up to one for each node. The learned parameters are the weights of the encoder $E$ and the weights of the decoder $\Delta$, which are annotated with red.}
	\label{fig:complete_algo}
\end{figure*}
\section{node2coords}\label{sec:node2barycoords}
We present now node2coords, an unsupervised graph representation learning algorithm that relies on the graph Wasserstein barycenter representation method, introduced in Section \ref{sec:barycentric_layer}. The proposed autoencoder architecture is shown in Fig. (\ref{fig:complete_algo}). The input of the encoder are the connectivity descriptors of the nodes $Z_n$, which capture their local structure. The node connectivity descriptors are passed through a linear layer followed by a softmax activation in order to obtain a small set of graph patterns $M_S$. In the decoder we employ Wasserstein barycenters as demonstrated in Section \ref{sec:barycentric_layer}, to reconstruct the node connectivity descriptors as Wasserstein barycenters of the graph patterns in $M_S$. Thus, node2coords learns both the graph patterns $M_S$ and the barycentric coordinates $\Lambda$. We give more details of each block of node2coords below.

\textbf{Input:}
Given the adjacency matrix $\mathcal{A}$ of a graph of $N$ nodes, we choose to define the matrix of node connectivity descriptors $Z_n$ as:
\begin{equation}
Z_n(i,j) = \frac{\tilde{\mathcal{A}}^n(i,j)}{\sum_{j=1}^N (\tilde{\mathcal{A}}^{n}(i,j))}\cdot
\end{equation} 
The matrix $\tilde{\mathcal{A}}$ is defined as $\tilde{\mathcal{A}}=\mathcal{A}+\alpha \mathbf{I}_N$, where $ \mathbf{I}_N$ is the $N$-dimensional identity matrix and $\alpha \in \{0, 1\}$. The matrix $\tilde{\mathcal{A}}^n$ is therefore computed as $\tilde{\mathcal{A}}^n=(\mathcal{A}+\alpha \mathbf{I}_N)^n$. The $i$-th row $Z_n(i, \cdot)$ is the connectivity descriptor of node $i$ and its support indicates the nodes that can be reached from node $i$ in up to $n$ hops. The value of the parameter $\alpha$ is set to $\alpha=0$ for $n=1$ because for $n=1$ the only nodes that can be reached from a node are its one-hop neighbors. However, for $n \geq 2$ the value of $\alpha$ is set to $\alpha = 1$. The reason for this choice is that for $n \geq 2$, a node can always reach itself by hoping to its first-hop neighbors and back. The value of the parameter $n$ depends on the size of the graph $G$. Specifically, for larger values of the number of nodes $N$, larger values for $n$ are required. An example of a connectivity descriptor for $n=1$ plotted on the graph is shown in Fig. (\ref{fig:connect_hist}). 

\textbf{Encoder:}
In the encoder, the node connectivity descriptors $Z_n$ are passed through a linear layer with $N \times S$ ($S<N$) parameters $E$ followed by a softmax activation. The $N \times S$ matrix $M_S$ obtained at the output of the encoder is therefore: 
\begin{equation}
M_S = \operatorname{softmax}(Z_nE),
\end{equation}
where the softmax activation of an $N$-dimensional vector $x_i$ is defined as:
\begin{equation}
\operatorname{softmax}(x_i)=\frac{e^{x_i}}{\sum_{j=1}^N e^{x_j}}\cdot
\end{equation}
The $S$ graph patterns in $M_S$ capture the most important structural properties of the graph and, therefore, we refer to them as \emph{graph structural patterns}. 

\textbf{Decoder:}
In the decoder, the node connectivity descriptors in $Z_n$ are reconstructed as Wasserstein barycenters of the $S$ graph structural patterns in $M_S$ using Wasserstein barycenters, as introduced in Algorithm \ref{algo:BarycenterLayer}. We obtain the optimal barycenter approximations $\hat{Z}_n$ as Wasserstein barycenters of $M_S$ by learning their barycentric coordinates $\Lambda$. 

In order to guarantee that the barycentric coordinates of each of the barycenter representations sum up to one, we introduce a change of variable, through a softmax activation, so that the barycentric coordinates $\Lambda(i, \cdot)$ of the barycenter approximation of the $i$-th node connectivity descriptor $Z_n(i, \cdot)$ are reparametrized through a matrix $\Delta$ as:
\begin{equation}\label{eq:cov}
\lambda_{i, k}=\frac{e^{\delta_{i, k}}}{\sum_{j=1}^{S}e^{\delta_{i, j}}}\cdot
\end{equation}

The node connectivity descriptors in $Z_n$ that are being reconstructed at the decoder are localized in the $n$-hop neighborhood of the nodes. Therefore, their barycenter approximations in $\hat{Z}_n$ are also localized in the $n$-hop neighborhood of the nodes. The localization of the barycenters in $\hat{Z}_n$ in $n$ hops, leads to learning patterns in $M_S$ that are localized in up to $n$-hops. As the entropy regularization parameter $\epsilon$ decreases, the graph structural patterns in $M_S$ tend to be localized in exactly $n$ hops. On the contrary, as $\epsilon$ increases the patterns in $M_S$ become more localized. \color{black} We note, however, that it is not possible to control the nodes  on which the graph structural patterns will have their highest values. \color{black} Furthermore, in the Wasserstein barycentric layer of the decoder the graph is taken into account through the diffusion distance cost $C=D_{\tau}^{p=1}$. The nature of the displacement interpolation obtained with Wasserstein barycenters \cite{Simou:2019} and the use of the diffusion distance cost, which captures the geometry of the underlying graph, leads to a small set of graph patterns in $M_S$ that highlight its structural properties.  

It is important to observe that the decoding step can be thought of as an embedding of the nodes in the space spanned by the graph patterns in $M_S$. The $i$-th node of the graph, as described by its connectivity $Z_n(i, \cdot)$, is embedded in the $S$-dimensional space defined by the patterns in $M_S$ by learning its $S$-dimensional coordinates $\Lambda(i, \cdot)$. As a result, each element of the $S$-dimensional embedding $\Lambda(i, \cdot)$ of the node $i$ quantifies the proximity, in terms of Wasserstein distance on the graph, of its connectivity descriptor to the $S$ graph structural patterns. \color{black} The dimensionality of the embedding space $S$ is a design choice and typically it depends on the number of clusters in the graph. \color{black}

\textbf{Optimization:}
We train node2coords in order to learn the graph representations $M_S$ and $\Lambda$ by minimizing a loss $\mathcal{L}(\hat{Z}_n, Z_n)$ between the node connectivity descriptors $Z_n$ and their reconstruction as barycenters $\hat{Z}_n$. In order to ensure that the reconstruction of each node connectivity descriptor in $Z_n$ is taken equally into account, we consider the normalized reconstruction loss 
$\mathcal{L}(\hat{Z}_n, Z_n)=\frac{\|Z_n - \hat{Z}_n\|^2_F}{\|Z_n\|^2_F}\cdot$

It can be seen from Algorithm \ref{algo:BarycenterLayer} and Eq. (\ref{eq:cov}) that $\mathcal{L}(\hat{Z}_n, Z_n)$ is differentiable with respect to $\Delta$ and $M_S$. Therefore, $\mathcal{L}(\hat{Z}_n, Z_n)$ is differentiable with respect to $\Delta$ and $E$. The minimization of the loss function is a non-convex problem:
\begin{equation}\label{eq:opt}
\operatorname*{min}_{E, \Delta} \frac{\|Z_n - \hat{Z}_n\|^2_F}{\|Z_n\|^2_F},
\end{equation}
that can be optimized with automatic differentiation \cite{paszke2017automatic} and stochastic gradient descent (SGD) \cite{lecun2012efficient}. \color{black} The number of barycenters computed in parallel $J$, introduced in Section \ref{ssec:efficient}, determines the batch size used for training with SGD and it is a design choice. As, the energy function in Eq. (\ref{eq:opt}) is non-convex, the graph structural patterns in $M_S$, and as a result also the barycentric coordinates $\Lambda$, will not be exactly the same for each run of node2coords. 
Finally, we note that larger values of the entropy regularization $\epsilon$, constitute the energy function in Eq. (\ref{eq:opt}) less non-convex. \color{black} 

\section{Experimental Results}\label{sec:experiments}
\subsection{Settings}
In this section, we evaluate the quality of the representations learned with node2coords for community detection and node classification tasks and examine their stability to perturbations of the graph structure. We compare the performance of our algorithm against the following unsupervised learning methods:
\begin{itemize}
	\item
	Laplacian Eigenmaps (LE) \cite{belkin2002laplacian}: A shallow-embedding method that finds $S$-dimensional node embeddings by keeping the eigenvectors of the graph Laplacian matrix that correspond to the $S$ smallest eigenvalues. LE embeddings naturally emphasize the clusters in the graph.
	\item
	DeepWalk \cite{perozzi2014deepwalk}: An algorithm that uses random walks on graphs to learn $S$-dimensional representations of nodes with a skip-gram model.
	\item
	node2vec \cite{grover2016node2vec}: A skip-gram method that uses biased random walks on graphs allowing for a trade-off between homophily and structural equivalence of the obtained node embeddings.
	\item
	SDNE \cite{wang2016structural}: An autoencoder that learns $S$-dimensional node embeddings at the ouput of the $N \times S$ linear layer of the encoder. SDNE embeddings preserve first-order and second-order node proximities.
	\color{black}
	\item
	DVNE \cite{zhu2018deep}: An autoencoder that learns $S$-dimensional Gaussian distributions in the Wasserstein space as the latent representation of the nodes. DVNE embeddings preserve the graph structure while simultaneously modelling the uncertainty of nodes.  
\end{itemize}

The above methods were chosen in order to ensure comparison with benchmark methods as well as state-of-the-art methods for unsupervised learning of graph representations without node features. 

We consider the following datasets:
\begin{itemize}
	\item
	Karate: The Zachary Karate network \cite{zachary1977information} is composed of $N=34$ nodes. Each node is a member of a Karate university club, which is split into two communities. 
	\item
	PolBooks: This dataset consists of a network of $N=105$ books about US politics published around the time of the 2004 presidential election and sold by the online bookseller Amazon.com \cite{Senn:2009}. Edges between books represent frequent copurchasing of books by the same buyers. The books belong to one of three classes ``liberal", ``conservative", ``neutral". 
	\item
	Citeseer4: Citeseer \cite{nr} is a dataset that consists of a citation network extracted from the Citeseer digital library. Nodes are publications and an edge exists between two nodes if either publication has cited the other one. The publications belong to one of 6 classes, where each class corresponds to a research area. Citeseer4 is a network of $N=1532$ nodes and corresponds to the giant component of the network obtained by the publications that belong to the 4 research areas of the Citeseer network.
\end{itemize}

\begin{figure*}[!b]
	\centering
	\includegraphics[width=\textwidth]{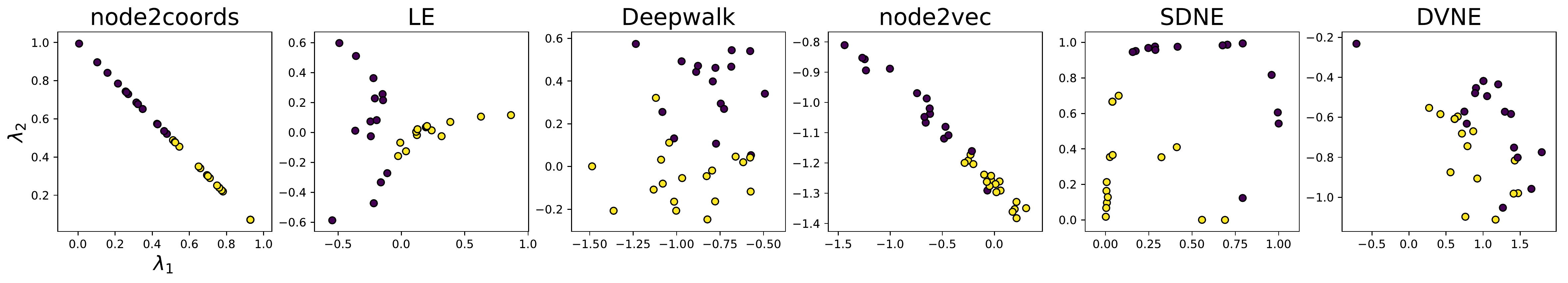}
	\caption{Embeddings obtained for the Zachary Karate network with node2coords, LE, Deepwalk, node2vec, SDNE and DVNE. For node2coords the two axes correspond to the barycentric coordinates $\lambda_1, \lambda_2$. The embeddings of the other methods are not in a known coordinate system. The embeddings of the two communities are most clearly separated with node2coords.}\label{fig:Karate_embeddings}
\end{figure*}

\subsection{Community Detection}
We first evaluate the node embeddings learned by node2coords for the task of community detection on the  Zachary Karate network. We also take the opportunity to explain in detail the representations learned with node2coords and build intuitions on how to select optimally its parameters.

We run node2coords for $C=D_{\tau=1}$ and $n=1$ because the Karate network is a small graph and therefore 1-hop structural information is sufficient. We pick $S=2$ because there are two communities in the graph. For the remaining parameters, the values that led to the most clearly separable embeddings were $\epsilon=0.03$, $\rho=0.05$. The choice of the value for the parameter $\epsilon$ determines how localized the graph structural patterns will be and $\rho$ controls the mass relaxation allowed.  Furthermore, $L=500$ iterations were sufficient for the barycenter computation to converge. 
We train using SGD with learning rate $\mu=0.01$. 

\begin{figure}[!h]
	\centering
	\includegraphics[width=0.25\textwidth]{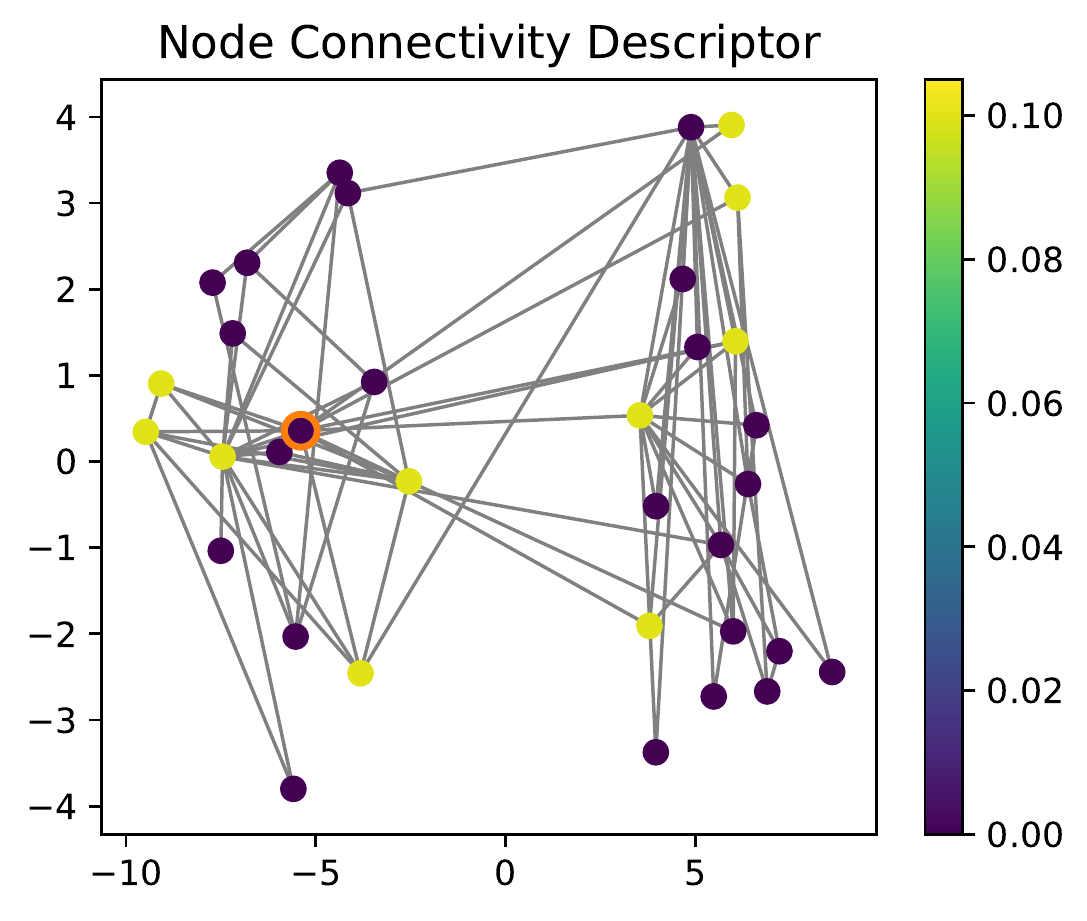}
	\caption{Connectivity descriptor of the node highlighted with the orange circle.}\label{fig:connect_hist}
\end{figure}

\begin{figure}[!h]
	\centering
	\includegraphics[width=0.5\textwidth]{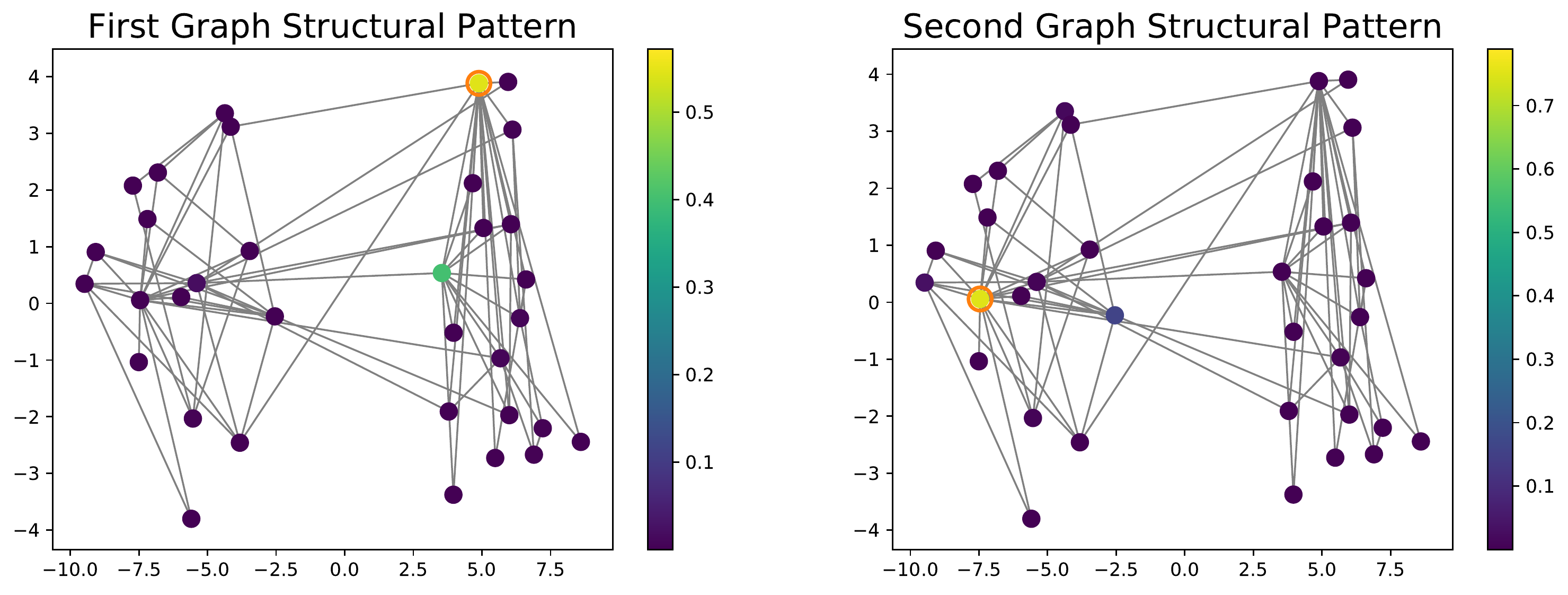}
	\caption{Graph structural patterns learned in $M_S$ for the Zachary Karate network. The colormap shows the range of intensities of the pattern on the graph. The patterns learned in $M_S$ are placed on each one of the communities.}\label{fig:Karate_DS}
\end{figure}

The graph structural patterns in $M_S$ are shown in Fig. (\ref{fig:Karate_DS}). It can be seen that the low-dimensional space $M_S$ comprises two very localized structural patterns which are placed on the two communities.  The embeddings obtained with node2coords are shown in Fig. (\ref{fig:Karate_embeddings}). The values of the barycentric coordinates $\lambda_1, \lambda_2$ of the nodes capture the proximity in terms of Wasserstein distance of the node connectivity descriptors relatively to the graph structural patterns of $M_S$. As an example, the connectivity descriptor  of the node in Fig. (\ref{fig:connect_hist}) is approximated in the decoder as $\operatorname*{argmin}_{u \in \Sigma_N}\sum_{i=1}^{2} \lambda_i W_{p}^{\epsilon, \rho}(M_S(\cdot, i),u)$ where $M_S(\cdot, 1), M_S(\cdot, 2)$ are the two  patterns shown in Fig. (\ref{fig:Karate_DS}) and $\lambda_1=0.43, \lambda_2=0.57$ are the barycentric coordinates learned for that node. Therefore, the nodes that are close to the first pattern in $M_S$ have a large $\lambda_1$ and a small $\lambda_2$, as can be seen by the embeddings of the nodes in yellow in Fig. (\ref{fig:Karate_embeddings}). The opposite is true for the nodes in the purple community. On the contrary, the node embeddings of the other methods lack interpretability as the value of the embedding of a node in each one of the two dimensions does not correspond to the proximity to a particular axes. 


We also show in Fig. (\ref{fig:Karate_embeddings}) the embeddings learned by Laplacian Eigenmaps, Deepwalk,  node2vec, SDNE and DVNE for latent dimensionality $S=2$ for the Zachary Karate Network. For node2vec, the parameters $p, q$ are set to $p=1, q=0.5$ which were proposed as optimal in \cite{grover2016node2vec} for the task of community detection where homophily among nodes is detected. For the remaining algorithms we optimize for the parameters in order to produce the most separable embeddings. For Deepwalk the number of walks is set to $\gamma=10$, the walk length to $t=10$ and the window size to $w=3$. For SDNE, the regularization for the first order proximities is equal to $\alpha=0.16$, the $L_2$ norm regularizer to avoid overfitting equal to $\nu=0.15$ and the reconstruction penalization parameter is set to $\beta=5$. \color{black} For DVNE we set the size of the hidden layer to $h=12$ and the parameter that controls the second-order proximity preservation to $\alpha_D=1$. \color{black} It can be seen that the embeddings obtained with all competitor unsupervised methods do not separate the two communities as clearly as node2coords. \color{black}

\subsection{Stability to Perturbations}\label{sec:generalization_perturbed}
First, we consider a stochastic block model (SBM) graph  \cite{holland1983stochastic} $G$ of $N=100$ nodes with probability of connection within the community equal to $p=0.4$ and probability of inter-community connection equal to $q=0.01$. We then consider perturbed versions $G^{\prime}$ of the graph $G$ by varying the probability $p$ within the range $p^{\prime}=\{0.15: 0.05 : 0.40\}$. Therefore, the perturbation affects the number of edges of the graph, but the number of nodes remains constant.

\begin{figure}[!h]
	\begin{subfigure}[b]{.49\textwidth}
		\includegraphics[width=\textwidth]{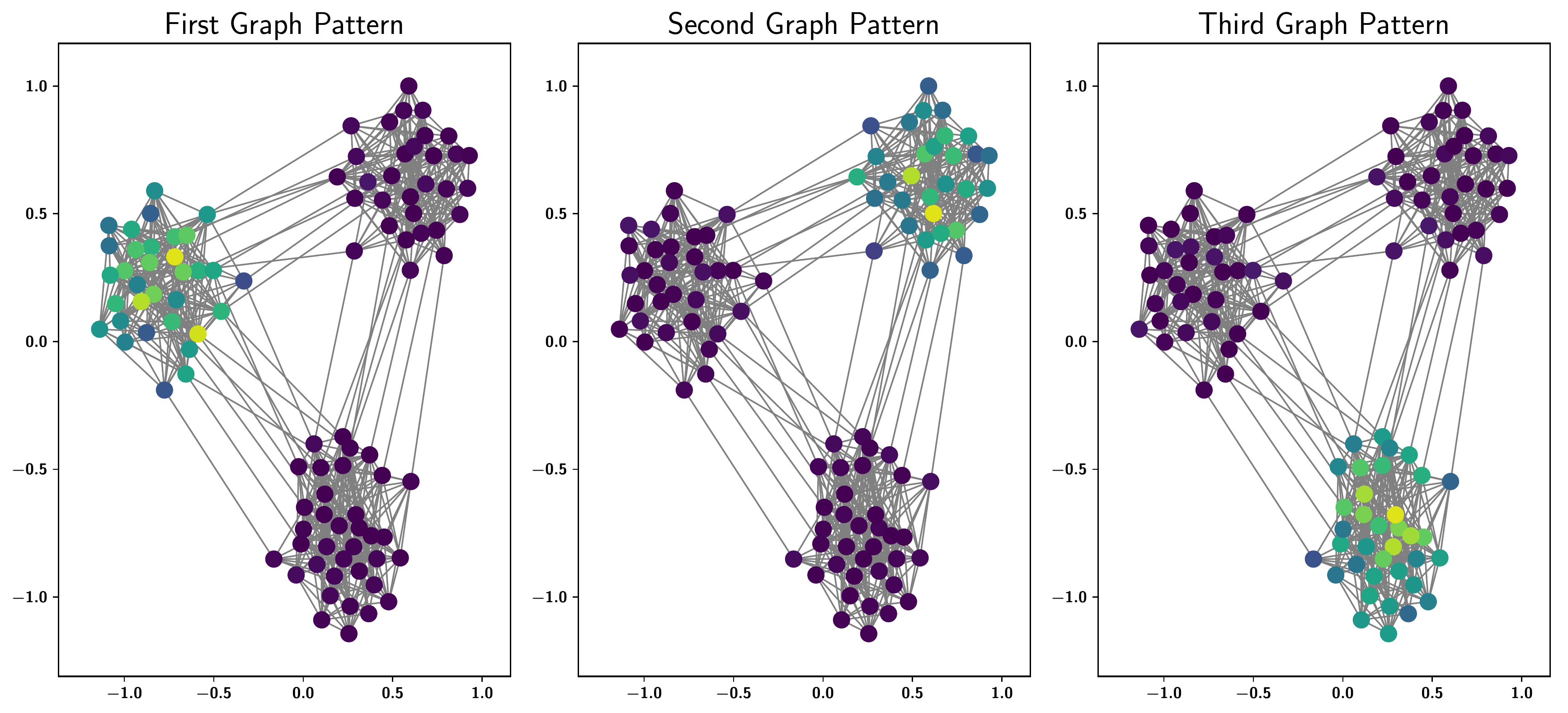}
		\caption{}
		\label{fig:SMB_clean}
	\end{subfigure}
	\hfill
	\begin{subfigure}[b]{.49\textwidth}
		\includegraphics[width=\textwidth]{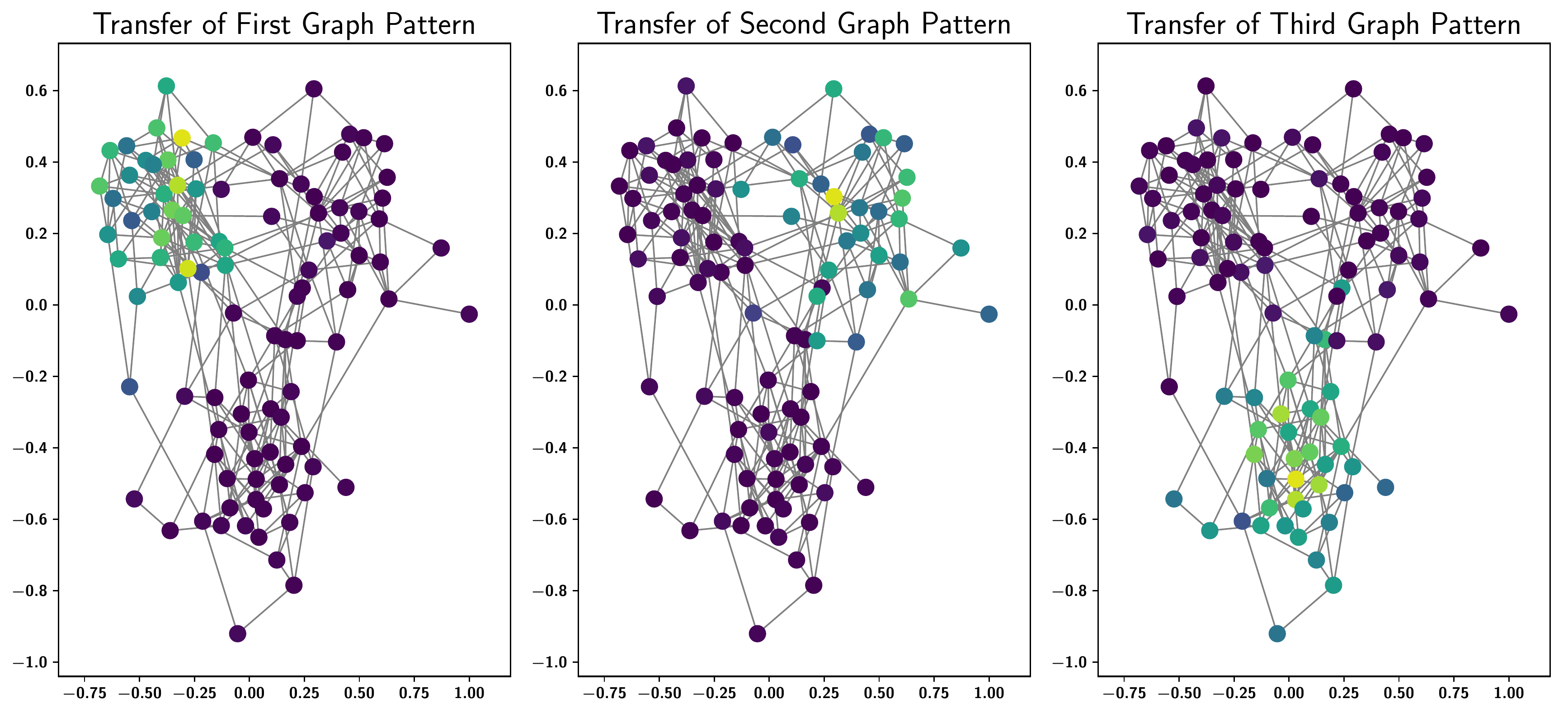}
		\caption{}
		\label{fig:SMB_pert}
	\end{subfigure}
	\caption{(a) Structural graph patterns of $M_S$ as learned for the graph $G$ with $p=0.4$. Each pattern identifies one of the communities. (b) Structural graph patterns of $M_S$ learned for the graph $G$ with $p=0.4$ transferred to the perturbed graph $G^{\prime}$ with $p^{\prime}=0.15$. The graph structural patterns remain meaningful for the perturbed graph $G^{\prime}$ as they clearly indicate the three communities.}
\end{figure}
\begin{figure}[!h]
	\centering
	\includegraphics[scale=0.18]{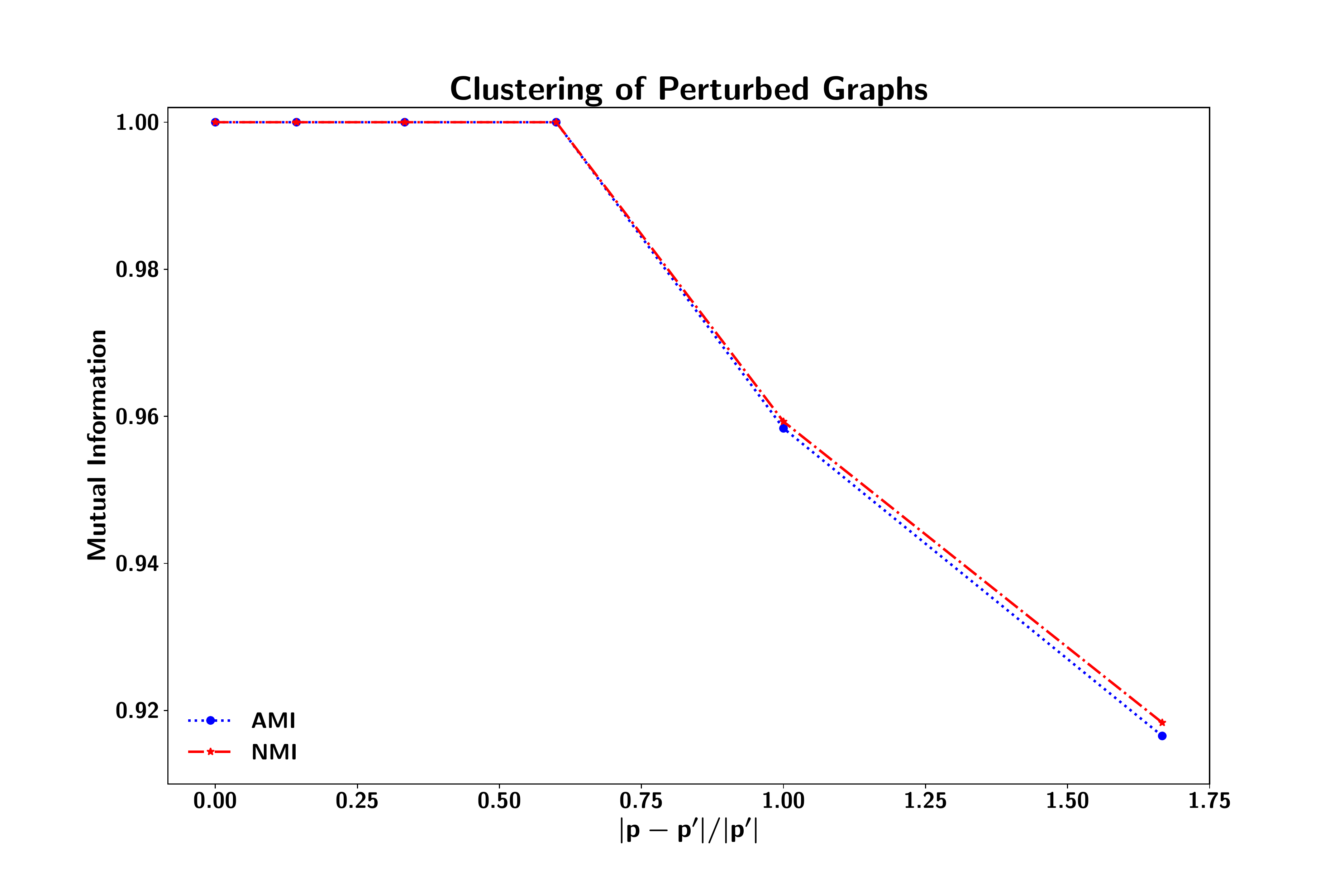} 
	\caption{AMI and NMI scores as a function of the relative change of the probability of connection within the community $\frac{|p-p^{\prime}|}{|p^{\prime}|}$.}
	\label{fig:SBM_perturb_clustering}
\end{figure}

We run node2coords with $n=1$, $S=3$, $\epsilon=0.01$ and $\rho=0.1$ and learn the space $M_S$ and the barycentric coordinates $\Lambda$ for the graph $G$. The graph structural patterns learned in $M_S$ for the clean graph $G$ are shown in Fig. (\ref{fig:SMB_clean}) and their transfer to the perturbed graph $G^{\prime}$ with $p^{\prime}=0.15$ in Fig. (\ref{fig:SMB_pert}). \color{black} The graph structural patterns are less localized in this case compared to those in Fig. (\ref{fig:Karate_DS}). This, as explained in Section  \ref{sec:node2barycoords}, is due to the fact that the entropy regularization parameter $\epsilon$ used for the SBM graph is smaller than the one used for the Karate network. However the interpretation of the graph structural patterns remains the same. \color{black} Specifically, it can be seen that the graph patterns in $M_S$ identify the three communities and thus they remain meaningful even when the actual graph changes. As a result, the perturbed graphs can be embedded in the low-dimensional space $M_S$ that was learned for the clean graph $G$. We confirm this intuition by evaluating the clustering result obtained using the node embeddings $\Lambda^{\prime}$ of the perturbed graphs in the space $M_S$ learned for the original graph $G$. For the perturbed graphs $G^{\prime}$ we only compute the barycentric coordinates $\Lambda^{\prime}$ of their nodes in the space $M_S$ learned on $G$. 
We apply $k$-means clustering to the barycentric coordinates of the nodes $\Lambda^{\prime}$ with $k=3$ and we compute the adjusted mutual information (AMI) and the normalized mutual information (NMI) \cite{vinh2010information} for the clustering result. The obtained AMI, NMI for the different perturbations are shown in Fig. (\ref{fig:SBM_perturb_clustering}). It can be seen that both the AMI and NMI are high even for large perturbations. Thus, we confirm that the perturbed graphs $G^{\prime}$ can be embedded in a meaningful way in the space $M_S$ learned for the clean graph $G$.

\begin{figure*}[!btp]
	\centering
	\includegraphics[width=\textwidth]{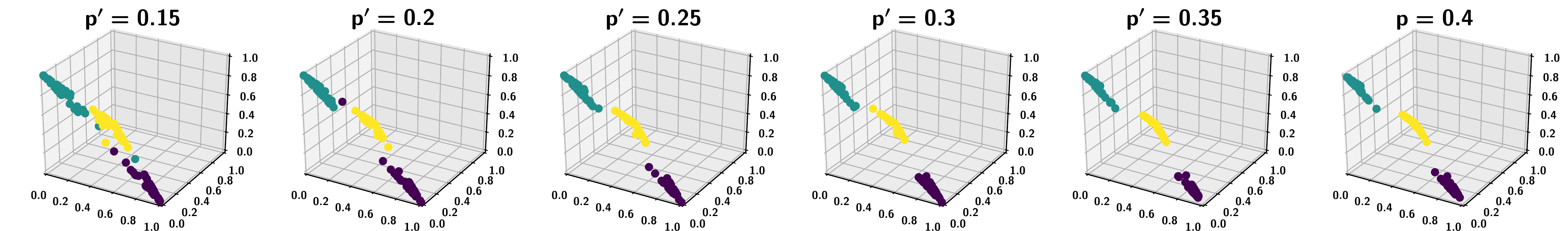}
	\caption{Embeddings of node2coords of the perturbed graphs $G^{\prime}$ in the space $M_S$ learned for the clean graph $G$.}\label{fig:SBM_node2barycoords_emb_stab}
\end{figure*}

\begin{figure}[!h]
	\centering
	\includegraphics[scale=0.18]{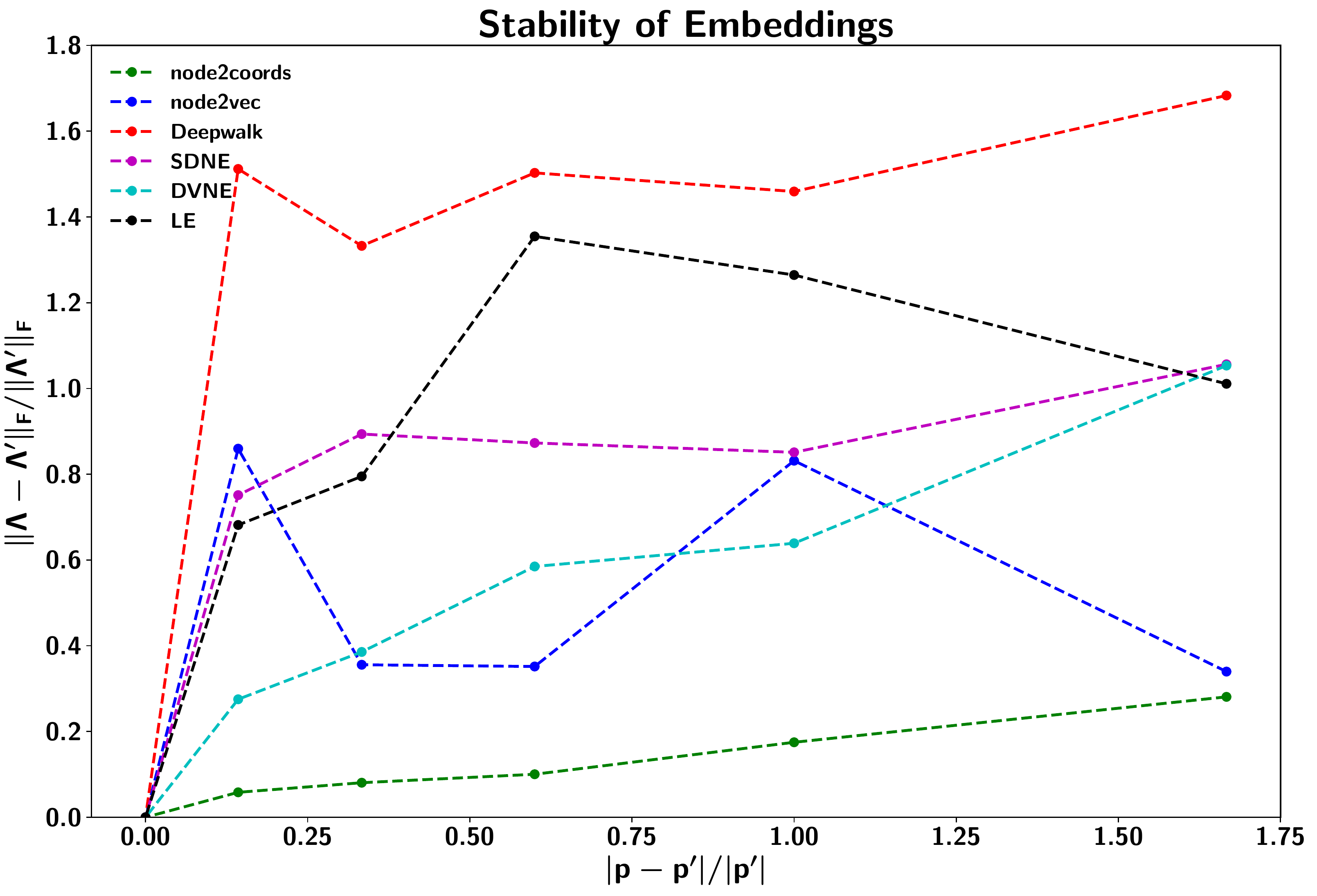}
	\caption{Relative change of the embeddings $\frac{\|\Lambda-\Lambda^{\prime}\|_F}{\|\Lambda^{\prime}\|_F}$ as a function of the relative change of the probability of connection within the community $\frac{|p-p^{\prime}|}{|p^{\prime}|}$ for the SBM graphs.}\label{fig:SBM_emb_stab}
\end{figure}

We further evaluate the relative change $\frac{\|\Lambda-\Lambda^{\prime}\|_F}{\|\Lambda^{\prime}\|_F}$ in terms of Frobenius norm of the barycentric coordinates $\Lambda^{\prime}$ of the perturbed graphs $G^{\prime}$ in comparison to the barycentric coordinates $\Lambda$ of the original graph $G$. In Fig. (\ref{fig:SBM_emb_stab}) we show the relative change of the embeddings obtained as a function of the relative change of the probability of connection within the community $\frac{|p-p^{\prime}|}{|p^{\prime}|}$. Laplacian Eigenmaps,  node2vec, Deepwalk, SDNE and DVNE do not learn a low-dimensional space as node2coords and, therefore, the only way to obtain the node embeddings of the perturbed graphs is by re-running the algorithms. It can be seen clearly that the embeddings obtained with node2coords are stable. \color{black} DVNE produces also relatively stable embeddings. This is expected as possible uncertainties of the node embeddings are accounted for through the variance of the Gaussian distribution. We notice also that the change in the node embeddings of SDNE seems to follow an increasing trend as the relative change of the probability of connection within the community increases. \color{black} The relative change in the embeddings of node2coords seems to increase linearly with the relative change in the probability of intra-connection $p$. This is also clearly seen in Fig. (\ref{fig:SBM_node2barycoords_emb_stab}) where we plot the node embeddings for node2coords. It can be seen that node embeddings obtained with node2coords change progressively as the value of the probability of connection $p$ changes. 

\begin{figure}[!h]
	\centering
	\includegraphics[scale=0.18]{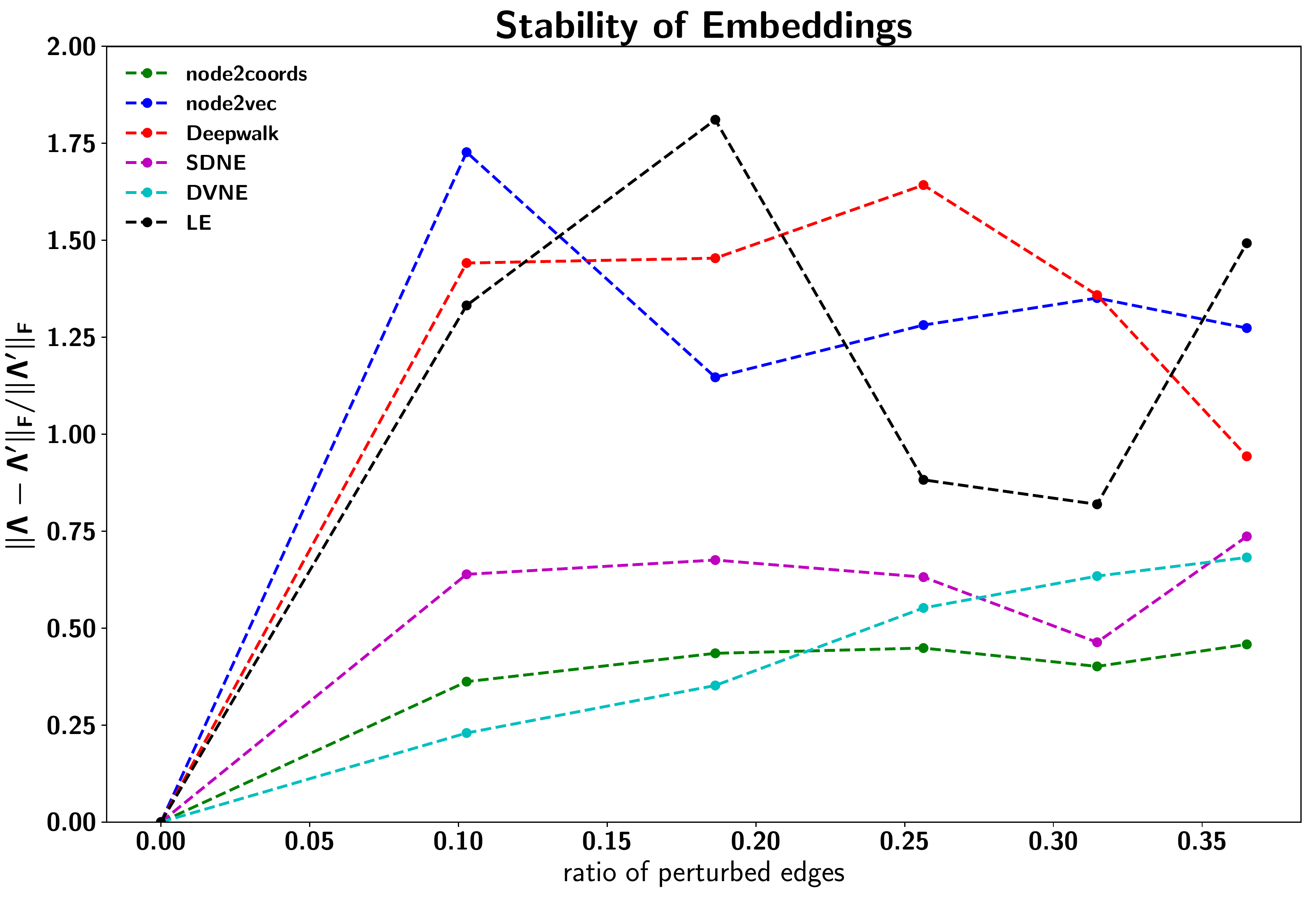}
	\caption{Relative change of the embeddings $\frac{\|\Lambda-\Lambda^{\prime}\|_F}{\|\Lambda^{\prime}\|_F}$ as a function of the percentage of perturbed edges in the graph $p=\frac{|\mathcal{E}_p|}{|\mathcal{E}|+|\mathcal{E}_p|}$ for PolBooks.}\label{fig:polbooks_emb_stab}
\end{figure}

\color{black}
Furthermore, we consider the graph $G$ of the PolBooks dataset and we create perturbed graphs $G^{\prime}$ by randomly adding edges. We denote the number of added edges as $|\mathcal{E}_p|$.  The maximal number of edges that can be added to this network until it becomes fully connected is $|\mathcal{E}_p|^{max}=N^2-|\mathcal{E}|=10143$. We define the ratio of added edges as $r_p=\frac{|\mathcal{E}_p|}{|\mathcal{E}_p|^{max}}$ and generate the perturbed graphs $G^{\prime}$ by varying the ratio $r_p$ in the range $r_p=\{0.01: 0.01: 0.05\}$. We run node2coords with $n=1$, $S=3$, $\epsilon=0.01$ and $\rho=0.1$ and learn the space $M_S$ and the barycentric coordinates $\Lambda$ for the graph $G$. Next, we compute the barycentric coordinates $\Lambda^{\prime}$ of the perturbed graphs $G^{\prime}$ in the space $M_S$  and evaluate the relative change of the barycentric coordinates $\frac{\|\Lambda-\Lambda^{\prime}\|_F}{\|\Lambda^{\prime}\|_F}$ in terms of Frobenius norm. In Fig. (\ref{fig:polbooks_emb_stab}) we show the relative change of the embeddings obtained as a function of the percentage of perturbed edges in the graph $p=\frac{|\mathcal{E}_p|}{|\mathcal{E}|+|\mathcal{E}_p|}$. It can be seen that again node2coords provides the most stable node embeddings followed by DVNE. We notice also that SDNE produces relatively stable embeddings. Furthermore, we notice that for the smaller perturbations $p$, DVNE produces the most stable embeddings. Contrary to the perturbations generated for the SBM graphs, where only the probability of connection within the community is perturbed, the perturbations created in this experiment add randomly edges either within or between clusters. Therefore, in this case the perturbed edges affect more strongly the clusters in the graph. These perturbations are more challenging for node2coords, since the geometry-aware cost relies on the diffusion distance, which naturally emphasizes the clusters in the graph. As the perturbation becomes stronger, the embeddings of DVNE become less stable than those of node2coords.

To conclude, we have shown experimentally that node2coords learns stable node embeddings. The advantage of the stability of the embeddings with node2coords is due to the fact that the low-dimensional space $M_S$ permits a registration of the nodes in the case of perturbed graphs.

\color{black}

\subsection{Node Classification}\label{sec:node_classification}

We now evaluate the features learned in an unsupervised manner with node2coords in the context of node classification tasks. The node embeddings learned by node2coords and the competitor methods are input to a one-vs-rest logistic regression classifier with L2 regularization. 

We consider train-test partitions of the data varying from 20\% to 80\%. For each partition we create 10 random splits of the data to train and test and provide results averaged over the 10 splits. For the evaluation of the classification results we compute the Macro-F1 score. The F1 score of a class is the harmonic mean of precision and recall. The precision for a class is the number of true positives divided by the total number of elements labeled as belonging to the class and recall is the number of true positives divided by the total number of elements that belong to the class. The Macro-F1 score is an unweighted mean of the F1 scores of each class.

\begin{figure*}[!t]
	\hspace{-10mm}
	\centering 
	\begin{subfigure}[b]{0.33\columnwidth}
		\includegraphics[width=1\textwidth]{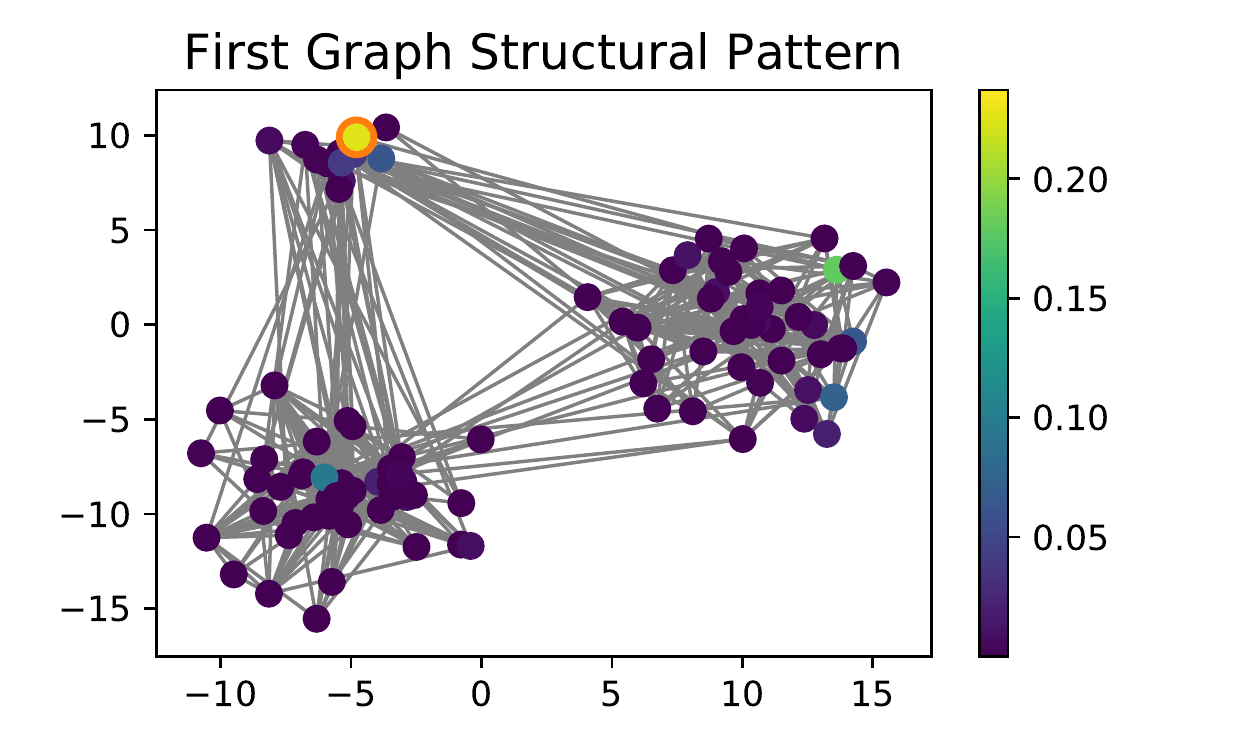}
		\label{fig:latent1}
	\end{subfigure}
	\begin{subfigure}[b]{0.33\columnwidth}
		\includegraphics[width=1\textwidth]{newestest_polbook_latent1.pdf}
		\label{fig:latent2}
	\end{subfigure}
	\begin{subfigure}[b]{0.33\columnwidth}
		\includegraphics[width=1\textwidth]{newestest_polbook_latent1.pdf}
		\label{fig:latent3}
	\end{subfigure}
	\hspace{-10mm}
	\vspace{-5 mm}
	\caption{Graph structural patterns learned for PolBooks. The node with the highest value of each pattern is highlighted with an orange circle. Each graph structural pattern indicates a cluster of the graph.}
	\label{fig:DS_polbooks}
\end{figure*}

We present in Table (\ref{tab:macroF1_polbooks}) the Macro-F1 scores for the PolBooks dataset and for train-test splits varying from 20\% to 80\%. 
For node2coords and the algorithms against which we evaluate its performance, the dimensionality of the embedding space is considered to be equal to $S=3$ as the network essentially has three clusters.
It can be seen that node2coords provides the highest Macro-F1 score for most training ratios. 
Indeed, even for the small latent dimensionality of $S=3$, it clearly identifies the small class ``neutral". This can be clearly seen also from the graph structural patterns of $M_S$ in Fig. (\ref{fig:DS_polbooks}). Specifically, DeepWalk, node2vec and SDNE make most of the classification errors for the small class ``neutral", which leads to reduced Macro-F1 scores. This shows that the embedding dimensionality $S=3$ is too small for these algorithms to properly capture the three clusters in the data, while it is sufficient for node2coords. \color{black}DVNE consistenly has the second best performance after node2coords when the  training ratio is more than 40\%.

\color{black}
In order to  illustrate the benefit of the barycentric layer compared to a  linear layer at the decoder, we also evaluate the performance of SDNE, which is composed of a linear layer in the encoder and the decoder, when  both its regularization parameters are set to $\alpha=0$ and $\nu=0$ and when its input are the node connectivity descriptors $Z_{n=1}$, which are also input at node2coords. We show in Table \ref{tab:macroF1_polbooks} the Macro-F1 scores obtained on the PolBooks dataset for SDNE with no regularizers (SDNE no reg.). It can be seen that node2coords significantly outperforms SDNE no reg., which illustrates the benefit of the barycentric layer compared to a  linear layer.
\color{black}

We now examine how node2coords performs on larger datasets. We show in Table \ref{tab:macroF1_citeseer3} the Macro-F1 scores for node classification on Citeseer4. The latent dimensionality for all methods is taken to be equal to $S=4$. 
Since Citeseer4 is a large graph we set $n=7$ and therefore the input of node2coords is $Z_{n=7}$. It can be seen again that node2coords provides the highest Macro-F1 scores for most train ratios. Therefore, node2coords is able to capture structural information for this larger graph. SDNE now has the second best performance, which shows that it is a method that scales well for large datasets. Also, node2vec performs consistently better than DeepWalk. This is due to the bias added to the random walks of node2vec, which leads to embeddings that capture both homophily as well as structural similarities of nodes. \color{black} The performance of DVNE drops significantly in this case. This is due to the low embedding dimensionality of $S=4$. Specifically, as observed by the authors in \cite{zhu2018deep}, the quality of the embeddings of DVNE decreases abruptly when the embedding dimensionality drops below a threshold. For instance, they report that for a graph of $N=2708$ nodes the minimum embedding dimensionality is $S=32$. Therefore, the embedding dimensionality of $S=4$ is insufficient for Citeseer4, which is a network of $N=1532$ nodes. \color{black} The worst performing method is LE, which does not scale well to larger graphs. The drop in performance of Laplacian Eigenmaps for large graphs could be due to the fact that it only takes into account first order proximities of the nodes. This is also indicated by the fact that the performance of SDNE, which takes into account second-order as well as first-order node proximities, improves for large graphs.

\color{black}
We also show in Table \ref{tab:macroF1_citeseer3} the Macro-F1 scores obtained on the Citeseer4 dataset for SDNE with no regularizers (SDNE no reg.) and input $Z_{n=7}$. It can be seen once again that node2coords significantly outperforms SDNE with no regularizers. This illustrates the benefit of the geometry-aware, non-linear interpolations of the barycentric layer, compared to linear interpolations.
\color{black}

\begin{table}[!h]
	\renewcommand{\arraystretch}{1.2}
	\caption{Macro-F1 score for node classification in PolBooks.}
	\label{tab:macroF1_polbooks}
	\centering
	\setlength\tabcolsep{4 pt}
	\begin{tabular}{c|ccccccc}
		\hline
		Train Ratio &20 \%  & 30 \% & 40 \% & 50 \%  & 60 \%  &  70 \% & 80 \% \\
		\hline
		LE &70.50 &72.83 & 75.47 & 70.14 &  72.85 &  77.27 & 73.33\\
		DeepWalk & 74.82 & \textbf{75.52} & 76.19 & 69.07 &  70.81 & 76.00 & \textbf{86.92} \\
		node2vec & 72.48 & \textbf{75.52} & 74.77 & 70.80 &73.61 &  80.76 &  \textbf{86.92}\\
		SDNE (no reg.) & 66.27 & 61.42 & 62.22 & 61.27 & 63.81 & 64.29 & 64.91 \\
		SDNE & 64.48 &  65.40 & 68.65 & 70.65 & 70.81 & 72.22 &79.16 \\
		DVNE &55.90 & 57.98 & 56.99 &  71.38 & 75.46 &  81.68 & 83.90\\
		\hline
		node2coords & \textbf{75.58} & 73.26 & \textbf{78.43} & \textbf{73.58} & \textbf{78.88} & \textbf{86.11} & \textbf{86.92}\\
		\hline
	\end{tabular}
\end{table}

\begin{table}[!h]
	\renewcommand{\arraystretch}{1.2}
	\caption{Macro-F1 score for node classification in Citeseer4.}
	\label{tab:macroF1_citeseer3}
	\centering
	\setlength\tabcolsep{4 pt}
	\begin{tabular}{c|ccccccc}
		\hline
		Train Ratio &20 \%  & 30 \% & 40 \% & 50 \%  & 60 \%  &  70 \% & 80 \% \\
		\hline
		LE & 30.78 & 32.47 & 31.81 & 32.87 & 34.09 & 34.57 & 36.63\\
		DeepWalk & 57.56 & 59.61 & 59.01 & 59.18 & 59.10 & 60.69 & 60.27 \\
		node2vec & \textbf{68.56} & \textbf{69.97} & 69.23 & 69.75 & 69.84 & 70.96 &74.44\\
		SDNE (no reg.) & 63.69 & 62.69 & 63.99 & 64.35 & 63.38 & 63.65 & 63.08 \\
		SDNE & 66.24 & 67.25 & 69.31 & 70.92 & 70.96 & 71.59 & 75.94\\
		DVNE & 38.48 & 40.60 & 40.24 & 40.63 & 40.86 & 39.47 & 41.45\\
		\hline
		node2coords & 66.94 & 68.63 & \textbf{70.58} &\textbf{72.14} & \textbf{74.12} & \textbf{75.97} & \textbf{78.80}\\
		\hline
	\end{tabular}
\end{table}

\subsection{Generalization to Unseen Nodes}
In this experiment we evaluate the ability of node2coords to generalize to nodes that have not been seen during the representation learning process. In order to do so, we use only a downsampled version of the adjacency matrix of the PolBooks network during the representation learning process. Specifically, we use a randomly selected set of nodes for training and subsample the $N \times N$ adjacency matrix $\mathcal{A}$ to obtain the $N^{train} \times N^{train}$ training adjacency matrix $\mathcal{A}^{train}$. From $\mathcal{A}^{train}$ we obtain the training connectivity matrix $Z_n^{train}$. With $Z_n^{train}$ as the input to node2coords, we learn the $N^{train} \times S$ space  $M_S^{train}$ and the $S$-dimensional barycentric coordinates for each of the training nodes in that space. We eventually use the $S$-dimensional barycentric coordinates to train a one-vs-rest logistic regression classifier with L2 regularization. 

As it has been shown above, the graph structural patterns learned with node2coords \color{black}for PolBooks \color{black} are sparse and they only have non-zero values on a small set of nodes within a given cluster. Therefore, we can upsample the patterns in $M_S^{train}$ by zero padding in order to obtain the $N \times S$ space  $M_S$. The graph structural patterns in $M_S$ obtained this way are meaningful as they indicate the clusters in the graph. We validate the quality of the patterns in $M_S$ by evaluating the classification performance of the unseen nodes' coordinates in the space defined by $M_S$. Specifically, we compute the barycentric coordinates of the unseen, test nodes in the space $M_S$ using the barycentric decoder of node2coords with fixed input $M_S$.  We predict their class labels using the trained logistic regression classifier and evaluate the classification accuracy. We consider downsampling partitions ranging from 50 \% to 90 \%.  For each partition we create 5 random splits of the data to train and test and provide results averaged over the 5 splits.

In Table (\ref{tab:acc_generalization_polbooks}) we show the classification accuracy for node2coords for downsampling ratios ranging from 50 \% to 90 \% (node2coords-DS) as well as the classification accuracy for the set-up of Section \ref{sec:node_classification} (node2coords). We can see that the algorithm generalizes well to nodes that were completely unseen while learning the representation $M_S$ with node2coords. When only 50 \% of the nodes are kept in $\mathcal{A}^{train}$ the classification accuracy for the nodes unseen during learning of $M_S$ is 76.15 \%.

Node2coords generalizes well to unseen nodes because the patterns learned for the downsampled graph capture the most important structural information which, in this case, corresponds to the clusters. This property is particularly useful in the case where the graphs under consideration are dynamic or temporally evolving. The ability of node2coords to learn such a meaningful low-dimensional representation of the graph, given only partial information of the graph structure, is unique to node2coords and cannot be reproduced by other methods for graph representation learning that only leverage structural information but not node features.

\begin{table}[!h]
	\renewcommand{\arraystretch}{1.2}
	\caption{Accuracy of node classification in PolBooks.}
	\label{tab:acc_generalization_polbooks}
	\centering
	\setlength\tabcolsep{4 pt}
	\begin{tabular}{c|ccccc}
		\hline
		Train Ratio & 50 \%  & 60 \%  &  70 \% & 80 \% & 90 \% \\
		\hline
		node2coords & 86.79 &  90.47 & 93.75 &95.23 & 100.00 \\
		node2coords-DS & 76.15 & 76.19 & 78.06 & 83.80 & 80.00  \\
		\hline
	\end{tabular}
\end{table}

\color{black}
\subsection{Parameter Sensitivity}
\begin{figure}[!t]
	\centering
	\includegraphics[scale=0.18]{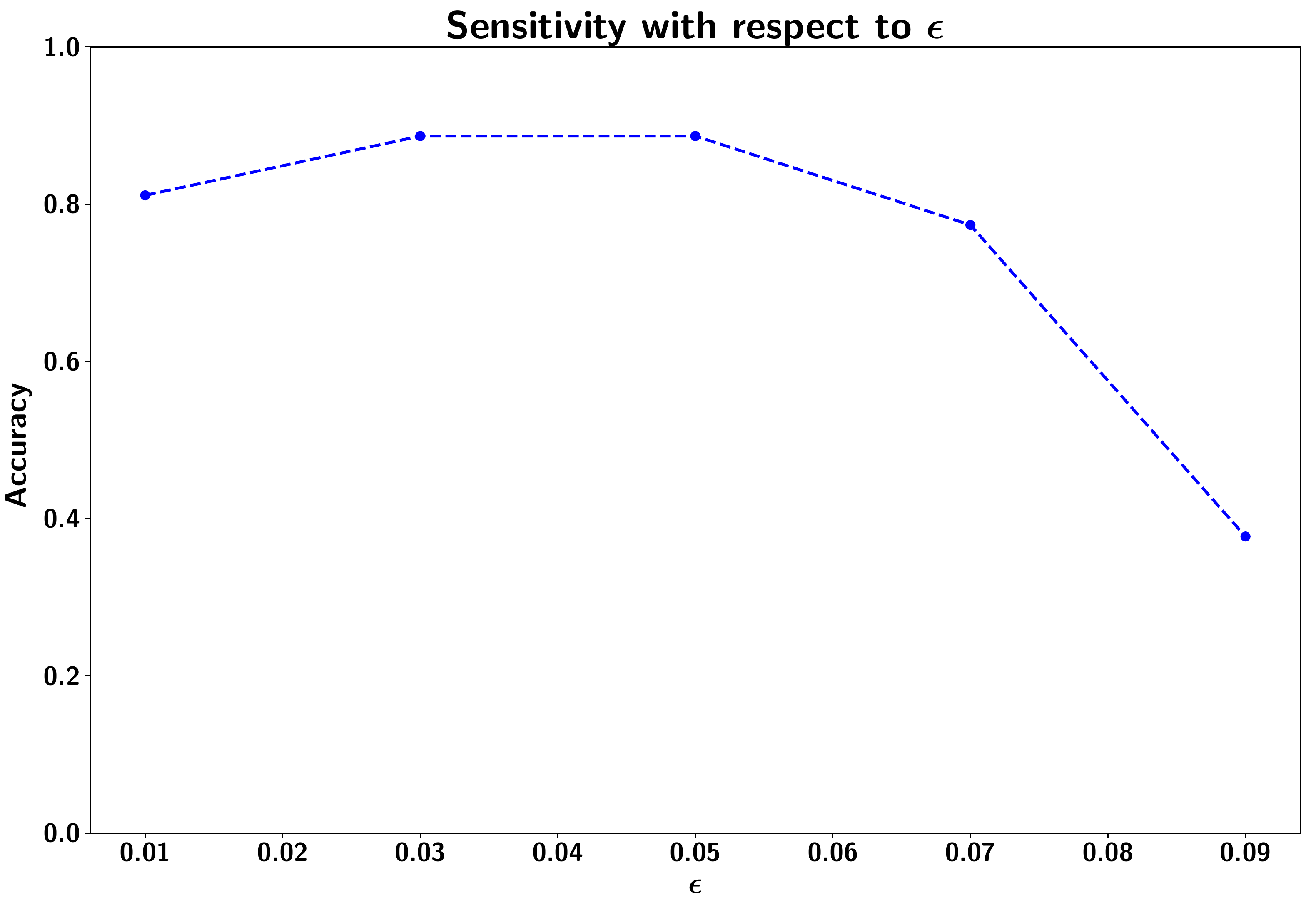}
	\caption{Sensitivity of the classification accuracy on the PolBooks dataset with respect to the entropy regularization parameter $\epsilon$.}\label{fig:epsilon_sensitivity}
\end{figure}

We investigate the sensitivity of the quality of the node embeddings learned with node2coords with respect to the parameter $\epsilon$. We set $\rho=0.1$ and vary $\epsilon$ in the range $\epsilon=\{0.01:0.01:0.09\}$. The classification accuracy as a function of $\epsilon$ is shown in Fig. (\ref{fig:epsilon_sensitivity}). It can be seen that the classification accuracy is relatively stable for $\epsilon$ in the range $\epsilon=\{0.01:0.01:0.05\}$ and drops for values of $\epsilon>0.05$. The reason for this drop in the perfromance for larger values of $\epsilon$ is directly linked with the quality of the graph structural patterns. As the entropy regularization parameter $\epsilon$ increases, the graph patterns are forced to be more localized. Significant increase of the regularization parameter $\epsilon$ forces the graph patterns to be Dirac $\delta$ functions, equal to 1 on a node and 0 everywhere else. In that extreme case, the barycentric coordinates will quantify the proximity with respect to $S$ nodes, \color{black}which is insufficient information for meaningful node embeddings. \color{black}

Further, we mention that the performance of node2coords is not significantly affected by the value of the parameter $\rho$. Specifically, even if $\rho$ is set to a smaller value than necessary, or equivalently if the mass preservation constraints are relaxed more than necessary, the mass of the barycenters will converge to that needed for the optimal reconstruction. We note that very large values of $\rho$ take us away from the ubalanced barycenter computation. In that case, there will be a drop in performance because the mass preservation constraints may penalize the optimal reconstruction of the node connectivity descriptors obtained with the Wasserstein barycenters. This is the reason why we have chosen to compute unbalanced Wasserstein barycenters in the decoder of node2coords.
\color{black}

\section{Conclusion}\label{sec:conclusion}
In this work we proposed node2coords, an autoencoder architecture with a novel Wasserstein barycentric decoder that learns low-dimensional graph representations without supervision. The proposed algorithm learns simultaneously i) a low dimensional space and ii) node embeddings that correspond to coordinates in that space. The low-dimensional space is defined by a small set of graph patterns that capture the most relevant structural information of the graph. The values of a node's embedding in that space can be interpreted as the proximity of its local connectivity to the corresponding graph patterns in terms of Wasserstein distance on the graph. 

We demonstrated how the low-dimensional space of node2coords can be used to obtain significantly more stable embeddings for slightly perturbed graphs, compared to other methods. Furthermore, we showed that the node embeddings of node2coords provide competitive or better results than those obtained with state-of-the-art methods for node classification tasks on real datasets. Finally, we confirm experimentally the ability to generalize to nodes that were unseen during the representation learning process, thus indicating the potential of node2coords to be used in dynamic settings. 
\bibliographystyle{IEEEtran}
\bibliography{IEEEabrv,new_node2coords_arxiv}


%








\end{document}